\documentclass[conference]{IEEEtran}
\usepackage{times}
\usepackage[pdftex]{graphicx}
\usepackage{fnpos}
\usepackage{upquote}
\usepackage{float}
\usepackage{verbatim}
\usepackage{tabularx}
\usepackage{fancyhdr}
\usepackage{mathtools}
\usepackage{pifont}
\usepackage{footmisc}
\usepackage{tablefootnote}
\usepackage{amsmath}
\usepackage{setspace}
\usepackage{amssymb}
\usepackage{dirtytalk}
\usepackage{graphicx}
\usepackage{lineno}
\usepackage{amsmath}
\usepackage{tablefootnote}
\usepackage{booktabs}
\usepackage{mathabx}
\usepackage{graphicx}
\usepackage{subcaption}
\usepackage[utf8]{inputenc}
\usepackage[export]{adjustbox}
\usepackage{wrapfig}
\usepackage{multirow}
\usepackage[table,xcdraw,x11names]{xcolor}
\usepackage{color, colortbl}
\definecolor{LightCyan}{rgb}{0.88,1,1}
\definecolor{lemonchiffon}{rgb}{1.0, 0.98, 0.8}
\usepackage{pgfplots}
\pgfplotsset{compat=newest}
\usepgfplotslibrary{groupplots}
\usepgfplotslibrary{dateplot}
\usepackage{algorithm}
\usepackage{algpseudocode}
\usepackage{anyfontsize}

\newcommand{\method}{POV-SLAM}
\DeclareMathOperator*{\argmax}{argmax}
\usepackage{makecell}
\usepackage[numbers]{natbib}
\usepackage{multicol}
\usepackage[bookmarks=true]{hyperref}
\hypersetup{
    colorlinks,
    linkcolor={red},
    citecolor={green},
    urlcolor={blue}
}
\usepackage[font={small}]{caption}

\algnewcommand{\LeftComment}[1]{\Statex \(\triangleright\) #1}
\newcommand{\customfootnotetext}[2]{{% Group to localize change to footnote
  \renewcommand{\thefootnote}{#1}% Update footnote counter representation
  \footnotetext[0]{#2}}}% Print footnote text

\begin{document}

% paper title
\title{POV-SLAM: Probabilistic Object-Aware Variational SLAM in Semi-Static Environments}

\author{\authorblockN{Jingxing Qian\textsuperscript{1},
Veronica Chatrath\textsuperscript{1,2},
James Servos\textsuperscript{3}, Aaron Mavrinac\textsuperscript{3}, Wolfram Burgard\textsuperscript{4}, \\
Steven L. Waslander\textsuperscript{1},
Angela P. Schoellig\textsuperscript{1,2}}
}

% \author{\authorblockN{Paper ID: 31}
% }

\maketitle

\customfootnotetext{1}{The University of Toronto Institute for Aerospace Studies and the University of Toronto Robotics Institute. \\Emails: \texttt{\{firstname.lastname\}@robotics.utias.utoronto.ca}}
\customfootnotetext{2}{The Technical University of Munich. \\Emails: \texttt{\{firstname.lastname\}@tum.de}}
\customfootnotetext{3}{Clearpath Robotics, Waterloo, Canada. \\
Emails: \texttt{\{jservos,amavrinac\}@clearpath.ai}}
\customfootnotetext{4}{The Technical University of Nuremberg. \\
Email: \texttt{wolfram.burgard@utn.de}}

\newcommand{\norm}[1]{\left\lVert#1\right\rVert}
\newcommand{\defeq}{\vcentcolon=}

\begin{abstract}

Simultaneous localization and mapping (SLAM) in slowly varying scenes is important for long-term robot task completion. Failing to detect scene changes may lead to inaccurate maps and, ultimately, lost robots. Classical SLAM algorithms assume static scenes, and recent works take dynamics into account, but require scene changes to be observed in consecutive frames. Semi-static scenes, wherein objects appear, disappear, or move slowly over time, are often overlooked, yet are critical for long-term operation. We propose an object-aware, factor-graph SLAM framework that tracks and reconstructs semi-static object-level changes. Our novel variational expectation-maximization strategy is used to optimize factor graphs involving a Gaussian-Uniform bimodal measurement likelihood for potentially-changing objects. We evaluate our approach alongside the state-of-the-art SLAM solutions in simulation and on our novel real-world SLAM dataset captured in a warehouse over four months. Our method improves the robustness of localization in the presence of semi-static changes, providing object-level reasoning about the scene.\customfootnotetext{5}{\label{fn_dataset}Dataset download and Supplementary Material are available at\\ \href{{https://github.com/Viky397/TorWICDataset}}{https://github.com/Viky397/TorWICDataset} \\ \indent  This work was supported by the Vector Institute for Artificial Intelligence in Toronto and the NSERC Canadian Robotics Network (NCRN).}

\end{abstract}

\IEEEpeerreviewmaketitle

\section{Introduction}

Simultaneous Localization and Mapping (SLAM) estimates a robot's pose within its environment, while at the same time creating a map of its surroundings. SLAM allows for autonomous navigation in GPS-denied situations, such as underground mines, office spaces, and warehouses. Many such tasks require robots to reliably repeat their trajectories over an extended period. However, most existing SLAM methods adopt the static world assumption~\cite{mur2015orb, grinvald2019volumetric, Rosinol20icra-Kimera, whelan2016elasticfusion} which typically does not hold in the real world, as scenes are subject to change from human or robot activity. For example, a scene may contain dynamic objects (e.g., forklift driving within a factory) and semi-static objects that change position over time (e.g., pallets, boxes). Lacking the ability to properly handle such changes might result in catastrophic failures such as corrupted maps, divergent pose estimations, and obstacle collisions. Such potential failures emphasize the importance of robust SLAM solutions in the presence of scene dynamics in order to achieve efficient and robust long-term robotic operation.

\begin{figure}[t!]
  \centering
  \includegraphics[width=1.0\columnwidth]{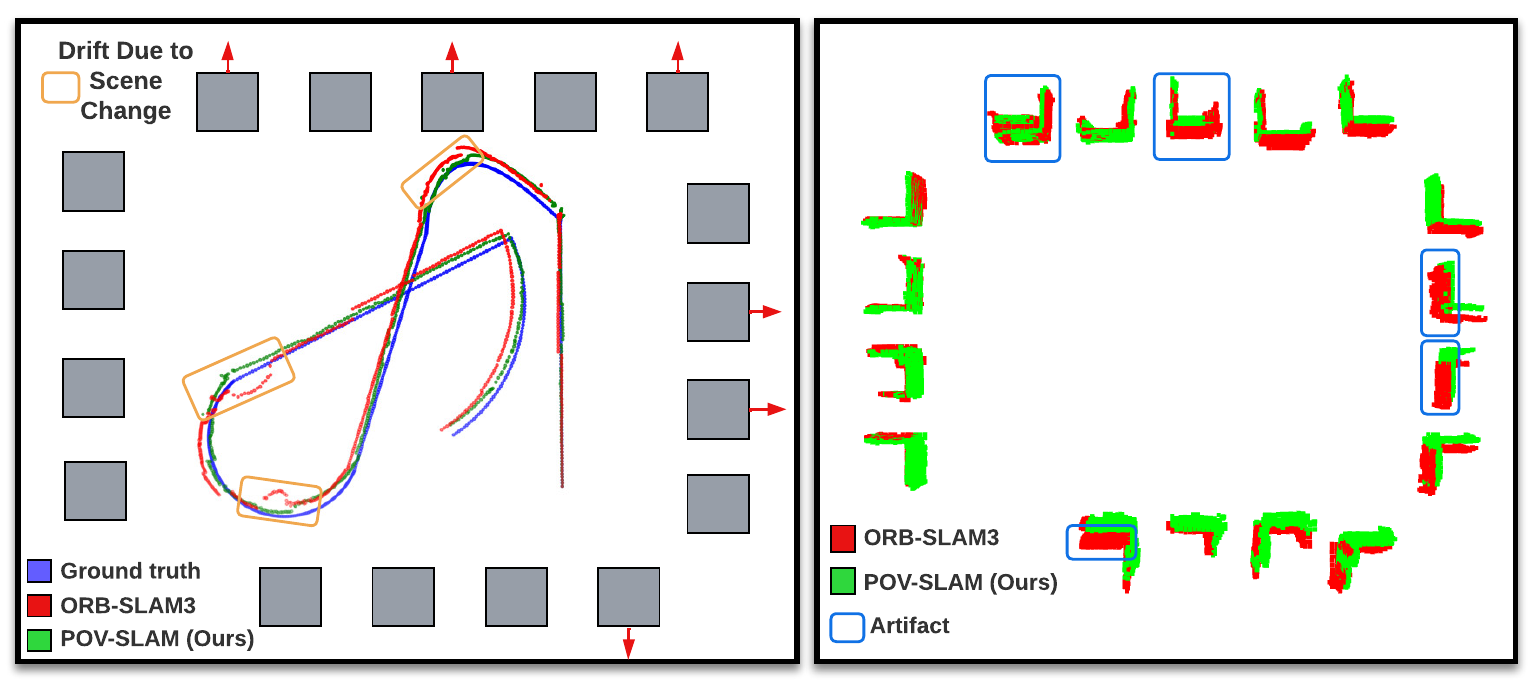}
  \caption{A qualitative comparison of the trajectory and scene reconstruction of a semi-static synthetic scene, BoxSim, where the boxes shift when not in the camera's field of view (indicated by the red arrows). Here we compare \method~to a state-of-the-art SLAM solution, ORB-SLAM3~\cite{campos2020orb}. As ORB-SLAM3 is a feature-based method, we feed its pose estimates into a semi-static mapping method, POCD~\cite{QianChatrathPOCD}, to generate the map. ORB-SLAM3 assumes a static world, suffering from localization drift (orange boxes, left) when changed objects are encountered, which leads to artifacts and incorrect map updates (blue boxes, right). Our approach explicitly infers object-level scene changes and provides more robust long-term localization and dense reconstruction. The reference desired reconstruction is shown in the top row of Figure \ref{fig:results_3dsim}.}
  \label{fig:fig1}
\end{figure}

Recent works have attempted to handle dynamic environments in one of two ways. The first strategy leverages semantic and geometric information to mask out all potentially dynamic objects, treating them as outliers~\cite{DOT, DS_SLAM, DM-SLAM, fusion++, Sun2019MovableObjectAwareVS, dsg}. Hence, the system only tracks against the static background, though it often covers a small portion of the sensor's field-of-view (FOV) in cluttered environments. The second strategy builds a model for each detected object. The system then either tracks the camera against the static background and refines the object models in a two-step pipeline, or performs camera and object tracking in a joint optimization problem~\cite{Rnz2018MaskFusionRR, Hachiuma2019DetectFusionDA, Xu2019MIDFusionOO, Wang2021DymSLAM4D}. However, the second strategy requires motion to be detected over consecutive frames, and long-term, semi-static changes where objects shift, disappear, or appear in the scene, have not been thoroughly studied in SLAM. Recent attempts to handle semi-static changes during map maintenance extend object-centric mapping methods to explicitly consider semi-static changes by estimating a consistency score for each object from a known robot pose~\cite{QianChatrathPOCD, visinsMulti, fehr2017tsdf, panoptictsdf}. Critically, when the robot pose is unknown, object consistency is difficult to calculate. The aforementioned consistency estimation methods can lead to multiple ambiguous and sub-optimal solutions. This limitation highlights the need for a statistically consistent method to infer both the robot pose and object consistency.

We tackle the challenge of simultaneous localization and object-level change detection in large semi-static scenes. We follow an object-aware strategy, as most mobile robots operate in environments consisting of rigid objects that move continuously or change location between visits. In addition to pose estimation, an up-to-date, object-level dense reconstruction is desired to provide rich geometric and semantic information for downstream tasks (e.g., perception-aware planning and control~\cite{doasican, swarm_microrobots}). We introduce a novel framework,~\method, which leverages recent works on object-level Bayesian consistency estimation for semi-static scenes~\cite{QianChatrathPOCD}, to tackle the challenge in a joint optimization problem. We derive a variational formulation to approximate the Gaussian-Uniform measurement model of potentially-changing objects, and use expectation-maximization (EM) to guarantee improvement of the evidence lower bound (ELBO) of a factor graph SLAM problem. At every EM iteration, the object consistencies and robot poses are refined using geometric and semantic measurements.

Additionally, there is a lack of SLAM datasets for long-term localization and mapping in large, semi-static environments. In collaboration with Clearpath Robotics, we extend the Toronto Warehouse Incremental Change Mapping Dataset (TorWIC), a long-term mapping dataset introduced in~\cite{QianChatrathPOCD}. Here, we present a real-world semi-static SLAM dataset in a warehouse with dynamic and semi-static changes that occur over four months\footref{fn_dataset}. To facilitate easier performance evaluation, we provide high-quality 3D scans of the entire warehouse and the ground truth robot trajectories, obtained from a Leica MultiStation and an onboard Ouster 128-beam LiDAR.     

Our proposed method is evaluated on: a 2D simulation to demonstrate the probabilistic framework in action and justify design choices, a synthetic semi-static dataset, and our real-world warehouse dataset. We analyze the reconstruction quality relative to a state-of-the-art (SOTA) dense semi-static mapping method~\cite{QianChatrathPOCD} and compare the localization accuracy against a SOTA feature-based SLAM method~\cite{campos2020orb} as well as a semi-static object-level SLAM~\cite{visinsMulti} approach. We show that our framework is robust to semi-static changes in the scene. The main contributions of our paper are: 

\begin{itemize}
    \item We derive a variational formulation for the Gaussian-Uniform bimodal measurement likelihood of potentially-changing objects. It exploits the Bayesian object consistency update rule introduced in~\cite{QianChatrathPOCD} and provides an evidence lower bound (ELBO) for efficient inference.
    
    \item We introduce an expectation-maximization (EM) algorithm to optimize factor graphs involving the variational measurement model for potentially-changing objects.

    \item We design \method, an object-aware, factor graph SLAM pipeline that tracks and reconstructs semi-static object-level changes. \method~builds on top of the SOTA SLAM \cite{campos2020orb} and semi-static mapping \cite{QianChatrathPOCD} methods, and uses our variational EM (VEM) strategy. The system is demonstrated both in simulation and in the real world.
    
    \item We release a new SLAM dataset captured in a warehouse over four months\footref{fn_dataset}. The environment contains static, semi-static, and dynamic objects as seen by RGB-D cameras and a 3D LiDAR. We also release a high-quality 3D scan of the warehouse and ground truth robot trajectories.
\end{itemize}

 In Section \ref{lit_review}, we review the SLAM methods for changing scenes. In Section \ref{sec:Pipeline_overview}, we present the key modules of the \method~pipeline. In Section \ref{sec:methodology}, we derive the variational measurement model and discuss the details of our VEM algorithm. Finally, we evaluate \method~in both simulated and real-world experiments in Section \ref{sec:results}. To the best of our knowledge, our method is the first to achieve joint localization and object-level change detection for large, semi-static environments.

%%%%%% Related Works
\section{Related Works} \label{lit_review}

\subsection{Visual SLAM}

Visual SLAM is a well-established type of SLAM, mainly achieved via either feature-based methods~\cite{mur2015orb,klein2007parallel,rgbdslam,vtar} or dense methods~\cite{newcombe2011kinectfusion, whelan2016elasticfusion}. Sparse methods match feature points of images, having lighter computational requirements, focusing on localization, whereas dense methods seek to construct accurate and more complete representations of the environment, useful for navigation and collision avoidance. 

In recent years, feature-based SLAM methods have gained traction for use with mobile robots in large environments, as they exhibit a high level of accuracy and efficiency. The seminal works of Mur-Artal \textit{et al.} in ORB-SLAM~\cite{mur2015orb} introduce a monocular, feature-based SLAM system with real-time camera relocalization. ORB-SLAM2~\cite{mur2017orb} and ORB-SLAM3~\cite{campos2020orb} extend~\cite{mur2015orb} with stereo and RGB-D information. ORB-SLAM3 remains a SOTA feature-based method~\cite{survey-slam, survey2-stereovis} and is extended to aid with our localization and map update strategy.

However, most current visual SLAM methods focus on static scenes, simply rejecting inconsistent landmarks from dynamic objects as outliers. As well, object-level scene information is ignored, resulting in inconsistent map updates when items move between robot passes. Our framework aims to use object-level understanding to track scene changes and aid with accurate localization in evolving scenes.

\subsection{Dynamic SLAM}

Dynamics and object-level reasoning in SLAM have been recently studied, and there exist two common strategies to handle changes. The first is to identify dynamics from input data, which can be extracted with a semantic segmentation network such as Mask R-CNN~\cite{maskrcnn},  discarding it completely \cite{DOT, DS_SLAM, DM-SLAM, fusion++, Sun2019MovableObjectAwareVS, dsg, SIVO}. Though this method is effective in the presence of a few dynamic objects, in cluttered environments, the static background is often only a small part of the sensor’s FOV and ignoring all dynamic objects could lead to an insufficient number of visual features for localization. 

The second strategy is to track the dynamic objects explicitly, which can be achieved using multi-object tracking (MOT)~\cite{Rnz2018MaskFusionRR,  Hachiuma2019DetectFusionDA, Xu2019MIDFusionOO, dynslam, strecke2019_emfusion}. DetectFusion~\cite{Hachiuma2019DetectFusionDA} uses semantic segmentation and motion consistency to extract both known and unknown objects. The work of Barsan \textit{et al.}~\cite{dynslam} uses instance-aware semantic segmentation and sparse scene flow to classify objects based on their activity. MID-Fusion~\cite{Xu2019MIDFusionOO} and EM-Fusion~\cite{strecke2019_emfusion} obtain object masks and construct a signed distance function (SDF) model for objects from depth information. Object poses are obtained by directly aligning depth measurements to their corresponding SDF models. VDO-SLAM~\cite{zhang2020vdoslam} and ClusterSLAM~\cite{clusterslam} group landmarks to form objects and exploit rigid body motion to construct a factor graph, jointly solving for robot and object poses. The aforementioned methods require scene changes to be observed in consecutive frames, rendering this strategy ineffective under changes that occur over a long time horizon. 

\subsection{Semi-Static SLAM}
SLAM in semi-static scenes is a difficult yet overlooked problem, that is crucial for long-term operation. One challenge in the presence of semi-static objects is ambiguity in the system state, caused by potential symmetry in the scene changes and the lack of continuously observed motion. 

Recent works on map maintenance involving semi-static objects all aim to estimate a consistency score, based on given robot poses, to determine which part of the map needs to be updated. Fehr~\textit{et al.}~\cite{fehr2017tsdf} update an SDF map by calculating voxel-level differences between signed distance functions of the stored map and incoming depth measurements. Schmid \textit{et al.}~\cite{panoptictsdf} maintain a set of object-level SDF sub-maps, propagating a stationarity score for each sub-map by calculating the overlap between their depth measurements and the existing map. Though intuitive, these overlap-based estimation methods are prone to localization errors. Gomez \textit{et al.}~\cite{object_pose_graph} model objects as cuboid bounding volumes and construct an object factor graph to estimate the object poses and their moveability scores in offline batch optimization. To obtain a more accurate and consistent object-level consistency score at runtime, Qian \textit{et al.} propose~\cite{QianChatrathPOCD}, a Bayesian update rule to iteratively propagate a probabilistic object state model using both geometric and semantic measurements, which was shown to be more robust against localization error. However, the aforementioned incremental mapping solutions all assume reliable robot poses are given.

Walcott \textit{et al.} propose a 2D LiDAR SLAM solution~\cite{walcott2012dynamic} that maintains a set of sub-maps for each region. The active sub-map is replaced with new measurements if there are inconsistencies, and then stacked to form the final map. Rosen \textit{et al.} incorporate a recursive Bayesian persistence filter~\cite{Rosen2016TowardsLF} into classic feature-based SLAM systems to estimate the consistency of each point feature. In a more recent work, Ren \textit{et al.}~\cite{visinsMulti} attempt to integrate object-level consistency estimation and 3D visual SLAM in the presence of semi-static and dynamic objects. The authors first perform dense visual SLAM using the static background to estimate the camera pose. They calculate the image-plane overlap between new object measurements and their previously mapped objects, reconstructing the object if the inconsistency is large. The visual features of unobserved mapped objects and new object observations are compared to perform association and relocalization. However, a known static background is required to track the camera motion, making this method a two-step process and rendering it unstable in the presence of a large number of potentially changing objects.

Rogers \textit{et al.}~\cite{SLAM-EM} and Xiang \textit{et al.}~\cite{RobustGraphSLAM} use an EM approach to handle semi-static point landmarks. The authors integrate the traditional landmark measurement model with a latent confidence score to weight its contribution in the cost function. The EM scheme is used to iteratively update the robot pose, landmark positions, and confidence scores. However, since the optimization process runs over the entire trajectory and rejection decisions are made based on a predefined threshold at the end of the process, these two methods are limited to offline settings. In our sliding window setup, landmark rejection decisions are revised probabilistically at every EM iteration during run-time, leading to more robust, fast, and accurate convergence. %EM-based algorithms have been applied to other components of SLAM systems, such as data association~\cite{Bowman2017ProbabilisticDA}.

\begin{figure*}[t!]
  \centering
  \includegraphics[width=\linewidth]{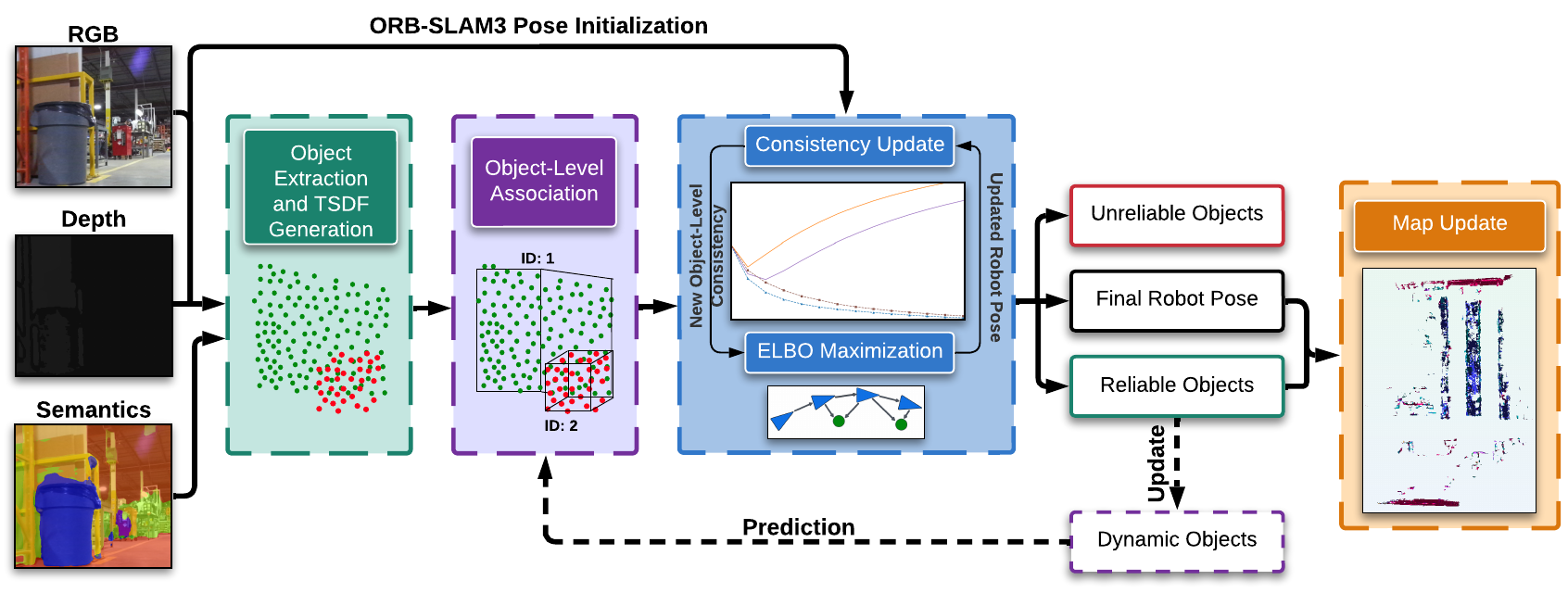}
    \caption{Our object-aware simultaneous localization and mapping framework (Section~\ref{sec:Pipeline_overview}) for semi-static environments. The system inputs are semantically annotated RGB-D frames (Section~\ref{pipeline_representations}). Object point clouds are first extracted and TSDFs are generated. Current object TSDFs are then associated with those stored in the object library (Section~\ref{sec:pipeline_extraction}). An object-level probabilistic consistency update and an evidence lower bound (ELBO) maximization (Section~\ref{sec:factor_graph_overview}) are performed iteratively to estimate the state of each object and localize the robot (Section~\ref{sec:iter_update_overview}). Upon convergence, the outputs are: unreliable objects, which are discarded, reliable objects, and the final robot pose, which are used to generate and update the dense scene map (Section~\ref{sec:obj_map_update}). The framework can be extended to additionally handle high dynamic objects (dotted lines) by following multi-object tracking (MOT) approaches~\cite{Hachiuma2019DetectFusionDA, zhang2020vdoslam}.}
  \label{fig:pipeline}
\end{figure*}

\section{System Description}
\label{sec:Pipeline_overview}

\subsection{Overview and Assumptions}
\label{sec:Overview_assumption}
This work focuses on long-term SLAM in the presence of semi-static objects. We aim to simultaneously localize the robot, and propagate a consistency estimate for each object. The robot localizes itself against objects with high measurement likelihoods, with changed objects being reconstructed once sufficient observations have been made. Finally, a truncated signed distance function (TSDF) map is produced to reflect the current scene configuration. 

The \method~system builds upon a recent semi-static map maintenance framework, POCD~\cite{QianChatrathPOCD}, and the SOTA feature-based RGB-D SLAM system, ORB-SLAM3~\cite{campos2020orb}. A flow diagram of our novel \method~system is shown in Figure~\ref{fig:pipeline}, which consists of five main stages. The following subsections provide an overview of each of the major components in the \method~pipeline.

We make the following assumptions in this work:

\begin{enumerate}
    \item The robot operates in a bounded indoor environment (e.g., warehouse or mall) where rigid objects are present.
    \item High-level prior knowledge of the objects is available, such as their semantic class, dimension, and likelihood of change.
    \item Objects can be added, removed, or shifted between robot traversals, though part of the environment should remain unchanged and observed by the robot.
    \item The robot starts its trajectory from a known pose.
\end{enumerate}

\subsection{SLAM Pipeline and Object Representation}
\label{pipeline_representations}
The \method~pipeline takes in a sequence of color and depth frames, $\mathcal{F}=\{\mathbf{F}_t\}_{t=1 \dots T}$, from a RGB-D camera, $\mathcal{C}$, as inputs at timestamps $t \in \{1 \dots T\}$. The pipeline outputs the 6-DoF world-to-camera transformations, $\mathcal{T}^{CW}=\{\mathbf{T}^{CW}_{t}=\{\mathbf{p}^{CW}_{t},\mathbf{q}^{CW}_{t}\}\}_{t=1 \dots T}$, with 3D position, $\mathbf{p}^{CW}_{t}$, and orientation, $\mathbf{q}^{CW}_{t}$, at each timestep $t$, along with a library of mapped objects, $\mathcal{O}=\{\mathbf{O}_i\}_{i=1 \dots I}$. Each object, $\mathbf{O}_i$, consists of: 
\begin{itemize}
    % added from POCD below
    \item a 4-DoF global pose $\mathbf{T}^{OW}_{i}$ with 3D position $\mathbf{p}^{OW}_{i}$ and heading $\phi^{OW}_{i}$, %\\$\mathbf{T}^{OW}_{i} = \{\mathbf{p}^{OW}_{i},\phi^{OW}_{i}\}$,
    \item a point cloud from accumulated depth data, $\mathbf{P}_{i}$, and the resulting TSDF reconstruction, $\mathbf{M}_i$, 
    \item a bounding box, $\mathbf{B}_i$, aligned with the major and minor axes of the object reconstruction,
    \item a semantic class, $c_i$,
    \item a state probability distribution, $p(l_i,v_i)$, to model the object-level geometric change, $l_i\in\mathbb{R}$, and the consistency, $v_i \in [0,1]$.
    \item a set of associated 3D landmark points in the world frame, $\mathcal{L}^{W}_i = \{ \mathbf{l}^{W}_{i,l} \in \mathbb{R}^3 \}_{l=1 \dots L}$,
    \item the relative positions of the landmark points with respect to the object pose, $\mathbf{T}^{OW}_{i}$, $\mathcal{L}^{O}_i = \{\mathbf{l}^{O}_{i,l} \in \mathbb{R}^3 \}_{l=1 \dots L}$,
\end{itemize}

% If an object is classified as dynamic (e.g., another mobile robot), a linear Kalman filter is created for the object, containing:
% \begin{itemize} 
%     \item a track history, $\hat{\mathcal{T}}^{OW}_{i,t} = \{\hat{\mathbf{T}}^{OW}_{i,1}\dots \hat{\mathbf{T}}^{OW}_{i,t}\}$,
%     \item an estimated pose covariance, $\hat{\mathbf{P}}_{i,t}$, at current time, $t$,
%     \item a process noise, $\mathbf{Q}_i$, and a measurement noise $\mathbf{R}_i$,
% \end{itemize}

As we consider indoor mobile robot applications, objects are restricted to only rotate around the $z$-axis, resulting in a 4-DoF pose, although extending to 6-DoF is trivial. The SLAM system is initialized with an empty object library, $\mathcal{O}=\varnothing$. Along with the camera pose and object models, the system also maintains a dense TSDF map which can be used for downstream tasks such as perception-based planning and control~\cite{doasican, swarm_microrobots}. 

\subsection{3D Observation Extraction and Data Association} \label{sec:pipeline_extraction}
When a new RGB-D frame, $\mathbf{F}_t$, is received by the system, a set of 3D observations, $\mathcal{Y}_{t} = \{\mathbf{Y}_{t,j}\}_{j=1 \dots J}$, is extracted and associated to the mapped objects by following the POCD semantic-geometric clustering and association strategy~\cite{QianChatrathPOCD}. Additionally, each observation, $\mathbf{Y}_{t,j}$, contains the unprojected 3D keypoints, $\mathcal{D}^{C}_{t,j} = \{\mathbf{d}^{C}_{t,j,d} \in \mathbb{R}^3 \}_{d=1 \dots D}$, detected from the masked color image. For each associated object-observation pair, $\{\mathbf{O}_i,\mathbf{Y}_{t,j}\}$, we also match the unprojected keypoints, $\mathcal{D}^{C}_{t,j}$, to the object landmark points, $\mathcal{L}^{C}_{i}$. %For objects labeled as dynamic, we first propagate their states using the estimated motion models to obtain the predicted poses, $\check{\mathbf{T}}^{OW}_{i,t}$, and covariance, $\check{\mathbf{P}}_{i,t}$, before performing association.

\subsection{Object Consistency-Augmented Factor Graph}
\label{sec:factor_graph_overview}

In POCD~\cite{QianChatrathPOCD} the authors introduced a Bayesian update rule to propagate an object-level state model, $p(l,v)$. This model consists of a Gaussian distribution which captures the magnitude of the object-level geometric change, $l$, and a Beta distribution which estimates the consistency between the incoming measurement and the previously mapped object, $v$. In this work, we exploit the Beta parametrization of consistency $p(v)\defeq\textrm{Beta}(v \mid \alpha, \beta)$, to estimate the reliability of the object observations in a factor graph optimization framework. 

We first consider a simple sparse SLAM problem in a semi-static scene, where previously mapped objects are either moved or unchanged when the robot revisits the region. Our goal is to estimate the robot trajectory, $\mathcal{T}^{CW}$, and determine which of the objects have changed. Existing methods such as ORB-SLAM3~\cite{campos2020orb} wrap landmark measurement residuals with a robust kernel (e.g., Cauchy loss function) and run optimization multiple times to reject outlier measurements. However, such approaches are not robust to large changes in the scene. Instead, similar to~\cite{SLAM-EM, RobustGraphSLAM}, we augment the joint likelihood of our sliding window estimation problem with the object-level Beta-parametrized consistencies, $\{p(v_i)\}_{i=1 \dots I}$, to explicitly model the reliability of each observed landmark:

\begin{subequations}
\label{eq:aug_model_orig}
\begin{align}
    &\log p(\mathcal{O}, \{\mathbf{T}^{CW}_t\}_{t=T-m \dots T},  \{\mathcal{Y}_t\}_{t=T-m \dots T})  \\
    & \quad \propto  \sum_{t} \log p(\mathbf{e}^{\textrm{pose}}_{t}) \label{eq:aug_model_orig_1} \\
    & \quad + \sum_{i} \sum_{l} \log p(\mathbf{e}^{\textrm{rigid}}_{i,l}) \label{eq:aug_model_orig_2}\\
    & \quad + \sum_{i} \sum_{l} \log p(\mathbf{e}^{\textrm{prior}}_{i,l}) \label{eq:aug_model_orig_3} \\
    & \quad + \sum_{t} \sum_{j} \sum_{d} \log p(\mathbf{e}^{\textrm{key-pt}}_{t,j,d}, \boldsymbol{\alpha}, \boldsymbol{\beta}) \label{eq:aug_model_orig_4}
\end{align}
\end{subequations}

The factor in Equation~\eqref{eq:aug_model_orig_1} is the transition model. We use the ORB-SLAM3 RGB-D front-end to obtain a visual odometry (VO) measurement in the body frame, $\mathbf{T}^{C}_{t-1,t}$, which is used as a prior to initialize the augmented factor graph: 

\begin{equation}
\label{eq:aug_model_pose}
\begin{aligned}
    p(\mathbf{e}^{\textrm{pose}}_{t}) &= \mathcal{N}(\mathbf{e}^{\textrm{pose}}_{t} \mid \mathbf{0}, \sigma^{2}_{\textrm{pose}}\mathbf{I})
\end{aligned}
\end{equation}

\noindent where $\mathbf{e}^{\textrm{pose}}$ is the stacked translation and rotation of the deviation $\mathbf{T}^{\textrm{offset}}_{t}$ between the estimated relative pose in the body frame and the VO measurement:

\begin{equation}
\label{eq:pose_prior_os3}
\begin{aligned}
    \mathbf{T}^{\textrm{offset}}_{t} &= (\mathbf{T}^{CW }_{t-1} \mathbf{T}^{CW -1}_t)^{-1} \mathbf{T}^{C}_{t-1,t} \\
\end{aligned}
\end{equation}

\noindent In practice, we find that this factor improves the stability of the nonlinear optimization.

The factor in Equation~\eqref{eq:aug_model_orig_2} constrains the relative positions of associated object landmarks with respect to the object frame to penalize the deformation of the object geometry:

\begin{equation}
\label{eq:aug_model_rigid}
\begin{aligned}
    p(\mathbf{e}^{\textrm{rigid}}_{i,l}) &= \mathcal{N}(\mathbf{e}^{\textrm{rigid}}_{i,l} \mid \mathbf{0}, \sigma^{2}_{\textrm{rigid}}\mathbf{I}) \\
    \mathbf{e}^{\textrm{rigid}}_{i,l} &= \mathbf{T}^{OW}_{i} \mathbf{l}^{W}_{i,l} - \mathbf{l}^{O}_{i,l} \\
\end{aligned}
\end{equation}

The factor in Equation~\eqref{eq:aug_model_orig_3} encourages landmark points to remain at their original positions during optimization. This is important, as objects that have changed but not been rejected can lead to localization errors and a corrupted map, especially at early stages of the optimization process:

\begin{equation}
\label{eq:aug_model_prior}
\begin{aligned}
    p(\mathbf{e}^{\textrm{prior}}_{i,l}) &= \mathcal{N}(\mathbf{e}^{\textrm{prior}}_{i,l} \mid \mathbf{0}, \sigma^{2}_{\textrm{prior}}\mathbf{I}) \\
    \mathbf{e}^{\textrm{prior}}_{i,l} &= \mathbf{l}^{W}_{i,l} - \mathbf{l}^{W}_{i,l,\textrm{prev}} \\
\end{aligned}
\end{equation}

\noindent Note that, for simplicity, we use Gaussian measurement likelihoods and isotropic covariance with magnitude $\sigma^{2}_{\textrm{pose}}, \sigma^{2}_{\textrm{rigid}}, \sigma^{2}_{\textrm{prior}}$ for these three factors. 

The factor in Equation~\eqref{eq:aug_model_orig_4}, the landmark measurement model between an object landmark point, $\mathbf{l}^{W}_{i,l}$ and its observation, $\mathbf{d}^{C}_{t,j,d}$, is more complicated, as a Gaussian likelihood is not sufficient to model possible changes in a semi-static scene. An intuitive approximation is to adopt the same Gaussian-Uniform mixture, weighted by the expectation of the Beta consistency model, $\mathbb{E}[v]$, as in~\cite{QianChatrathPOCD}:

\begin{equation}
\label{eq:aug_model_keypt_naive}
\begin{aligned}
    p(\mathbf{e}^{\textrm{key-pt}}_{t,j,d}) &= \mathbb{E}[v] \mathcal{N}(\mathbf{e}^{\textrm{key-pt}}_{t,j,d} \mid \mathbf{0}, \sigma^{2}_{\textrm{key-pt}}\mathbf{I}) \\
    \quad &+ (1-\mathbb{E}[v])\mathcal{U}(\lVert\mathbf{e}^{\textrm{key-pt}}_{t,j,d}\rVert_2 \mid 0,e_{\textrm{max}})\\
    \mathbf{e}^{\textrm{key-pt}}_{t,j,d} &= \mathbf{T}^{CW -1}_{t} \mathbf{d}^{C}_{t,j,d} - \mathbf{l}^{W}_{i,l} \\
\end{aligned}
\end{equation}

\begin{figure}[t!]
  \centering
  \includegraphics[width=0.92\linewidth]{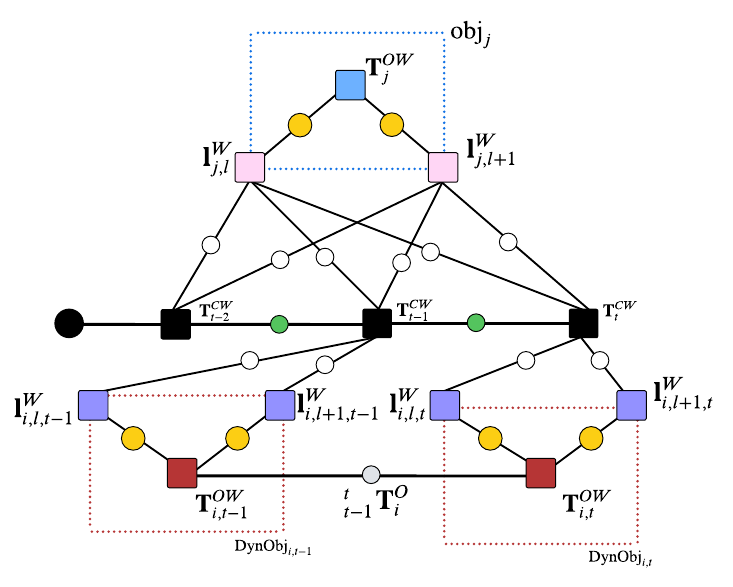}
    \caption{A factor graph representation of our probabilistic, object-aware SLAM method, solved in the M-step of each EM iteration (Section~\ref{sec:max_mix}), with \textbf{blue} semi-static and \textbf{red} optional dynamic objects. \textbf{Black squares}: robot poses at each time step, \textbf{black circle}: robot prior factor, \textbf{green circles}: odometry factors, \textbf{white circles}: ELBO of object landmark measurement factors (Section~\ref{sec:ELBO}), \textbf{pink} and \textbf{purple}: landmarks associated with static and dynamic objects, respectively, \textbf{orange circles}: object rigidity priors, \textbf{grey circle}: optional dynamic object velocity factor.}
  \label{fig:posegraph}
\end{figure}

\noindent This mixture model consists of two parts: 1) a zero-mean Gaussian component with an isotropic measurement covariance, $\sigma^{2}_{\textrm{key-pt}}$, for the unchanged scenario, and 2) a uniform component with a predefined maximum association distance, $e_{\textrm{max}}$, for the changed scenario in which the object could be anywhere. However, using the single point estimator, $\mathbb{E}[v]$, could lead to an inaccurate estimation as it does not capture the full Beta consistency distribution. We present a variational formulation to derive an ELBO for the landmark measurement model in Section~\ref{sec:ELBO}, which is efficient to implement, and shown to provide better convergence behavior than the single point approximation in Equation \eqref{eq:aug_model_keypt_naive}. Figure~\ref{fig:posegraph} illustrates the complete factor graph.

Optionally, our framework can be extended to handle dynamic objects in the scene. We follow the strategy in \cite{zhang2020vdoslam}, where the poses and associated landmarks of moving objects are modeled at each timestamp in the window, temporally constrained by the estimated velocity. This strategy is tested in simulation, as discussed in the Supplementary Material. 

\subsection{Iterative Object Consistency Update and Pose Estimation}
\label{sec:iter_update_overview}
The augmented optimization problem in Equation \eqref{eq:aug_model_orig} consists of both unknown parameters, which are the robot trajectory and the object poses with their associated landmark positions, and a set of unobserved latent variables, which are the object consistencies. Although the problem is complex to solve directly, a favoured approach to solving such estimation problems involving latent variables is iteratively via EM. We introduce an EM-based method in Section~\ref{sec:methodology}, which leverages our variational landmark measurement model to solve the factor graph iteratively at every frame, $\mathbf{F}_t$.

\subsection{Object and TSDF Map Update}
\label{sec:obj_map_update}

Once optimization is complete, we extract the new robot and object pose information, and update the map and object library. All new object observations not associated to the previous map are integrated into the TSDF map and added to the object library, $\mathcal{O}$. A large pseudo-change is used to penalize the consistency of objects currently in the camera frustum, but not associated with any observations. Objects \textit{accepted} by the VEM optimization, as discussed in Section \ref{sec:max_mix}, are considered consistent with the map, and their observations are integrated into the object's TSDF model, $M_i$. Their object state models are then propagated by one step based on the new robot pose estimate. The states of objects not accepted by the VEM optimization are also propagated by one step, though their observations are not integrated, as they are no longer consistent with their previous models. If an object's consistency expectation, $\mathbb{E}[v_i]$, falls below a pre-defined threshold, $\theta_{\textrm{consist}}$, the object is removed from the library, and all associated voxels in the TSDF model are reinitialized. Note that the rejected objects are not discarded immediately after optimization to ensure robustness against potential measurement noise and pose estimation error in the current frame. When dynamic objects are considered, we can update their motion models with their new pose estimates using a Kalman filter.

%%%%%% Methodology
\section{Methodology}
\label{sec:methodology}

In this section, we discuss the details of our VEM method: 1) In the E-step we compute the ELBO for the expectation of the landmark measurement likelihood for potentially changing objects, and 2) in the M-step we optimize the approximated factor graph to update the robot and object states. Algorithm~1 in the Supplementary Material outlines how our pipeline processes one frame to update the robot and object states.

\subsection{E-Step: ELBO of Measurement Likelihood for Potentially Semi-Static Objects}
\label{sec:ELBO}

In Section \ref{sec:factor_graph_overview}, the cost function for the augmented factor graph SLAM problem is introduced, where each object and its associated landmark points share a Beta-parametrized consistency estimate. Such a problem is challenging to optimize, even with the EM algorithm. Moreover, as discussed earlier, a single point approximation using the consistency expectation, $\mathbb{E}[v]$ (Equation \eqref{eq:aug_model_keypt_naive}), does not capture the full Beta consistency model. In this section, we focus on the E-Step of the VEM algorithm and derive the ELBO for the expectation of each object landmark's measurement likelihood in Equation~\eqref{eq:aug_model_orig}, based on the robot trajectory, object landmark position, and object consistency estimated in the previous EM iteration. 

Consider a single landmark from an object. At frame $T$ and EM iteration $n$, we obtain a Beta consistency posterior, $\textrm{Beta}(\alpha, \beta)$, for the object by following the Bayesian method introduced in~\cite{QianChatrathPOCD}, with respect to the current frame measurement, $\mathbf{d}^{C}_T$, the previous iteration's landmark position estimate, $\mathbf{l}^{W}$, and robot pose estimate, $\mathbf{T}^{CW}$. Note that these variables are treated as constants in the E-Step. The timestamps and indices in the notation are dropped for clarity. The object's true consistency, $\pi \in \{0,1\}$, can be considered as a sample from a Bernoulli distribution parametrized by $v$, with $\pi=1$ indicating the object has not changed. We can then write a generative process, $p(\pi, v) = p(\pi \mid v) p(v \mid \alpha, \beta)$, where:
\begin{equation} \label{eq:gen_process}
\begin{aligned}
v &\sim \textrm{Beta}(\alpha, \beta) \\
\pi &\sim \textrm{Bernoulli}(v) 
\end{aligned}
\end{equation}

Unchanged objects will follow a zero-mean, isotropic Gaussian measurement model and moved objects can be anywhere in the scene. The measurement residual, $\mathbf{e}_T$, is defined to be the 3D point-wise distance: 
\begin{equation} \label{eq:meas_residual}
\begin{aligned}
 \mathbf{e}_T &= \mathbf{T}^{CW -1} \mathbf{d}^{C}_T- \mathbf{l}^{W} \\
\end{aligned}
\end{equation}
We can then rewrite the Gaussian-Uniform measurement model weighted by the sampled object consistency, $\pi$, as:
\begin{equation} \label{eq:ours_meas_model}
\begin{aligned}
p(\mathbf{e}_T) & \defeq p(\mathbf{e}_T \mid \mathbf{T}^{CW },\mathbf{l}^{W},\pi) \\
 & = \mathcal{N}(\mathbf{e}_T \mid \mathbf{0},\sigma^2\mathbf{I})^{\pi}\mathcal{U}(\lVert \mathbf{e}_T \rVert_2 \mid 0, e_{\textrm{max}})^{1-\pi}
\end{aligned}
\end{equation}

Since $\pi$ is sampled from the generative process shown in Equation \eqref{eq:gen_process}, Equation \eqref{eq:ours_meas_model} involving dependent latent variables, $\boldsymbol{\omega} = \{\pi,  v \}$, is challenging to maximize. Fortunately, we can apply the mean field approximation~\cite{murphy2012machine} by assuming the two latent variables are fully independent, $p(\pi,v) \simeq q(\pi)q(v)$. This would allow us to write a variational lower bound, $\mathcal{L}$, for the evidence, $\log p(\mathbf{e}_T \mid \mathbf{T}^{CW },\mathbf{l}^{W},\alpha, \beta)$:

\begin{equation}
\label{eq:full_elbo}
\begin{aligned}
    &\mathcal{L}(\boldsymbol{\omega},\mathbf{T}^{CW },\mathbf{l}^{W}) =  \mathbb{E}_{q(\boldsymbol{\omega})} \left[ \log \frac{p(\mathbf{e}_T,\boldsymbol{\omega} \mid \mathbf{T}^{CW },\mathbf{l}^{W},\alpha, \beta)}{q(\boldsymbol{\omega})} \right]
\end{aligned}
\end{equation}

\noindent where the joint likelihood is

\begin{equation}
\begin{aligned}
&\log p(\mathbf{e}_T, \boldsymbol{\omega} \mid \mathbf{T}^{CW }, \mathbf{l}^{W}, \alpha, \beta) \\
& \quad = \log p(\mathbf{e}_T \mid \mathbf{T}^{CW }, \mathbf{l}^{W}, \pi) + \log p(\pi \mid v) + \log p(v \mid \alpha, \beta) \\
& \quad = \pi[\log v + \log \mathcal{N}(\mathbf{e}_T \mid \mathbf{0},\sigma^2\mathbf{I})]  \\
& \quad\quad +(1-\pi) [ \log(1-v)+ \log \mathcal{U}(\lVert \mathbf{e}_T \rVert_2 \mid 0, e_{\textrm{max}}) ]\\
& \quad\quad + \log \textrm{Beta}(v \mid \alpha, \beta) \\
\end{aligned}
\end{equation}

Following the mean field approximation, the optimal $q(\pi)$ and $q(v)$ that maximize the lower bound~\eqref{eq:full_elbo} are:

\begin{equation} \label{eq:q_pi_q_v}
\begin{aligned}
\log q(\pi)=& \pi \big[ \mathbb{E}[\log v]+ \log \mathcal{N}(\mathbf{e}_T\mid\mathbf{0},\sigma^2\mathbf{I})\big] \\
& +(1-\pi) \big[ \mathbb{E}[\log (1-v)] \\
 & \quad \quad + \log \mathcal{U}(\lVert \mathbf{e}_T \rVert_2 \mid 0, e_{\textrm{max}})\big] + const \\
\log q(v) =& \mathbb{E}[\pi] \log v + \mathbb{E}[1-\pi] \log (1-v) \\
 & +\log \textrm{Beta}(v \mid \alpha, \beta)+const
\end{aligned}
\end{equation}

Now, the expectation of the probability that the object did not change, $\mathbb{E}[\pi]$, can be computed based on the current measurement and estimates:

\begin{equation}
\begin{aligned}
&\mathbb{E}[\pi] = q(\pi=1) \\
&= \eta \exp \{\mathbb{E}[\log v] +\log \mathcal{N}(\mathbf{e}_T \mid \mathbf{0},\sigma^2\mathbf{I})\} \\
&\mathbb{E}[1-\pi] = q(\pi=0)\\
&= \eta \exp \{\mathbb{E}[\log (1-v)]  +\log \mathcal{U}(\lVert \mathbf{e}_T \rVert_2 \mid 0, e_{\textrm{max}})\} 
\end{aligned}
\end{equation}

\noindent Here, $\eta$ is a normalizing factor, and $\mathbb{E}[\log v]$ and $\mathbb{E}[\log (1-v)]$ can be computed from the property of the Beta distribution:

\begin{equation}
\begin{aligned}
&\mathbb{E}[\log v]=\psi(\alpha)-\psi(\alpha+\beta) \\
&\mathbb{E}[\log (1-v)]=\psi(\beta)-\psi(\alpha+\beta) \\
\end{aligned}
\end{equation}

\noindent where $\psi(\cdot)$ is the digamma function. Finally, we can compute the lower bound:

\begin{equation} \label{eq:elbo_full}
\begin{aligned}
&\mathcal{L}(v,\pi,\mathbf{T}^{CW },\mathbf{l}^{W}) \\
&=\mathbb{E}[\pi] \log \mathcal{N}(\mathbf{e}_T \mid \mathbf{0},\sigma^2\mathbf{I}) \\
&\quad +\mathbb{E}[1-\pi] \log \mathcal{U}(\lVert \mathbf{e}_T \rVert_2 \mid 0, e_{\textrm{max}}) + const
\end{aligned}
\end{equation}

Comparing to the naive approximation in Equation~\eqref{eq:aug_model_keypt_naive}, the ELBO is a mixture between a log-Gaussian mode and a log-Uniform mode. However, the new weights, $\mathbb{E}[\pi]$ and $\mathbb{E}[1-\pi]$, incorporate the full Beta consistency model as well as the likelihood of the two modes. This provides a more statistically consistent measurement model for potentially changing objects. We refer the reader to the Supplementary Material for a more detailed derivation, as well as a performance comparison against the single point approximation in Equation~\eqref{eq:aug_model_keypt_naive}.
    
\subsection{ELBO Tightness and Assumptions} \label{sec:tightness}

We repeat the ELBO estimation (Equation~\eqref{eq:elbo_full}) presented in Section~\ref{sec:ELBO} for all observed objects and their landmarks in the scene, which we substitute into the joint likelihood, discussed in Section~\ref{sec:factor_graph_overview}, to construct a lower bound to the original optimization cost (Equation~\eqref{eq:aug_model_orig}) for our sliding window SLAM problem. The new factor graph can be solved efficiently using an available SLAM solver, such as \textit{g2o} \cite{g20_pub}. 

Unfortunately, sub-optimal or diverged solutions are likely to occur. The mean field approximation used in the measurement ELBO tends to be overconfident~\cite{murphy2012machine}, especially when the Beta consistency estimate is uncertain. On the other hand, the ELBO tightens when the Beta distribution approaches a unit impulse, i.e., when $\alpha \gg \beta$ or $\alpha \ll \beta$. This implies that when object consistency estimates are uncertain, the lower bound can be improved but there is no guarantee to improve the true joint likelihood, as some moved objects can be misclassified as unchanged. Nonetheless, with additional iterations the ELBO tightens, improving the true likelihood. This convergence behavior requires that: 1) a good prior robot pose is available, and 2) some distinguishable, unchanged objects are observed by the robot. We believe these are reasonable assumptions to make in the semi-static SLAM problem. Most robots deployed in industrial settings depart from and return to pre-determined charging stations. Visual place recognition techniques can also be used to initialize the system. Moreover, if the robot only observes changed objects, then it is not possible to determine the global pose of the robot just using vision data. Without inertia or off-board anchor sensors (e.g., IMUs and UWBs), the system will converge to a minimum-cost state but there is no guarantee to the correctness. We provide simulation results in the Supplementary Material to illustrate the system's behavior under adversarial scenarios.

\subsection{M-Step: Factor Graph Optimization} \label{sec:max_mix}

In order to exploit the aforementioned assumptions and encourage the optimizer to make use of static objects with higher certainty to perform system updates, a max-mixture~\cite{Olson-max-mixture} approach is adopted to guide the optimization process. At every gradient descent step, for every object landmark, a weighted log measurement likelihood is computed for the unchanged and moved scenarios, and a decision is made on whether the measurement should be \textit{accepted} in computing the gradient: 

\begin{equation} \label{eq:max_mixture_1}
\begin{aligned}
m = \argmax \{  &\log \mathbb{E}[v] \mathcal{N}(\mathbf{e}_T \mid \mathbf{0},\sigma^2\mathbf{I}),  \\
& \log (1-\mathbb{E}[v])  \mathcal{U}( \lVert \mathbf{e}_T \rVert_2 \mid 0, e_{\textrm{max}})\}
\end{aligned}
\end{equation}

\begin{equation}  \label{eq:max_mixture_2}
\begin{aligned} 
&\Tilde{\mathcal{L}}(v, \pi, \mathbf{T}^{CW },\mathbf{l}^{W}) \\ 
&\defeq 
\begin{cases}
    \mathbb{E}[\pi] \log \mathcal{N}(\mathbf{e}_T \mid \mathbf{0},\sigma^2\mathbf{I}), & \text{if } m = 0\\
    \mathbb{E}[1-\pi] \log  \mathcal{U}(\lVert\mathbf{e}_T \rVert_2 \mid 0, e_{\textrm{max}}), & \text{if } m = 1
\end{cases}
\end{aligned}
\end{equation}

\noindent This approximation excludes objects with lower measurement likelihood from contributing to the overall cost, achieving faster and more accurate convergence when the ELBOs are not tight. Rejected objects are not deleted immediately, but revised at every gradient step. Note that we choose $\mathbb{E}[v]$ instead of $\mathbb{E}[\pi]$ to weight the measurement likelihoods when making the rejection decisions. Empirical results show that $\mathbb{E}[\pi]$, despite being a more accurate estimate, could be highly noisy due to the overconfidence in the mean field approximation. On the other hand, $\mathbb{E}[v]$ comes from the Bayesian update rule, thus providing smoother gradients to achieve more stable convergence. More details and ablation studies are provided in the Supplementary Material.

Substituting the approximated ELBO (Equation~\eqref{eq:max_mixture_2}) into Equation~\eqref{eq:aug_model_orig}, and maximizing the new factor graph, we obtain the updated robot and object states for the next EM iteration:

\begin{equation}
\label{eq:aug_model_final}
\begin{aligned}
\mathcal{O}, \mathcal{T}^{CW} &= \\
    &\argmax_{\mathcal{O}, \mathcal{T}^{CW}} \log p(\mathcal{O}, \mathcal{T}^{CW},  \{\mathcal{Y}_t\}_t)  \\
    & \quad \propto  \sum_{t} \log p(\mathbf{e}^{\textrm{pose}}_{t}) \\
    & \quad + \sum_{i} \sum_{l} \log p(\mathbf{e}^{\textrm{rigid}}_{i,l}) \\
    & \quad + \sum_{i} \sum_{l} \log p(\mathbf{e}^{\textrm{prior}}_{i,l}) \\
     & \quad + \sum_{t} \sum_{i} \sum_{l} \Tilde{\mathcal{L}}(v_i, \pi_i, \mathbf{T}^{CW}_t,\mathbf{l}^{W}_{i,l})
\end{aligned}
\end{equation}

Our VEM formulation ensures a monotonically increasing ELBO until a zero-gradient solution but does not guarantee convergence to an optimum. Global optimality is inherently challenging, but our descent method mostly provides high-quality solutions when assumptions in Section~\ref{sec:tightness} are met.

%%%%%% Experimental Results
\begin{figure*}[ht!]
  \centering
  \includegraphics[scale=.7]{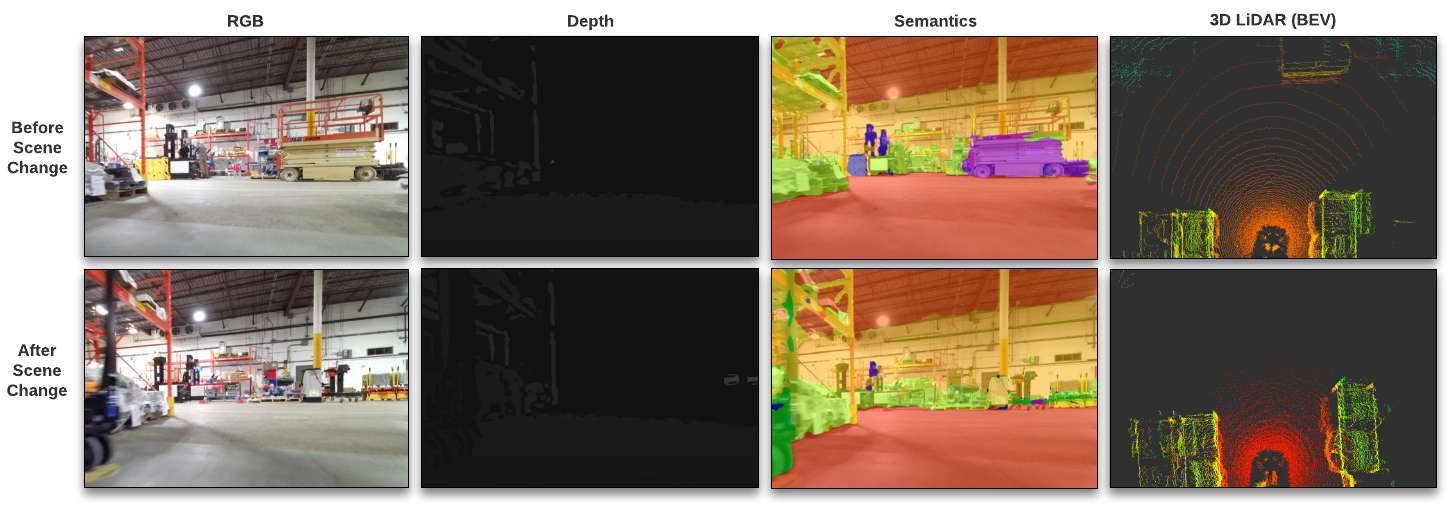}
  \caption{A sample RGBD, semantically labelled, and 3D LiDAR frame captured by the sensors before \textbf{(top)} and after \textbf{(bottom)} the scene changes. The forklift in the top frame, captured June 15, 2022, is no longer present in the bottom frame, captured Oct. 23, 2022.}
  \label{fig:scene_change}
\end{figure*}

\section{Experimental Results}
\label{sec:results}
\subsection{Experimental Setup}
We verify the performance of our framework qualitatively and quantitatively by comparing both the map reconstruction and robot trajectory error of \method~to:
\begin{itemize}
    \item ORB-SLAM3~\cite{campos2020orb}: A SOTA sparse visual SLAM solution, which assumes the world is static.
    \item VI-MID~\cite{visinsMulti}: A recent object-level SLAM method for small (5m$\times$5m) semi-static scenes. The method performs dense RGB-D tracking on certainly-static regions based on semantics for camera localization, before updating the object states. As the code was unavailable, we modify ORB-SLAM3 to exclude features from potentially changing objects during pose estimation, and use POCD~\cite{QianChatrathPOCD} for object change detection and mapping. Our custom implementation is referred to as \say{Ours-MID}.
\end{itemize}

In essence, ORB-SLAM3 uses all features, Ours-MID employs certainly-static features, and \method~probabilistically selects features from likely-unchanged objects. In a static scene, \method~should revert to regular batch SLAM where only the Gaussian mode of the measurement model is active. To demonstrate the capabilities of \method, we evaluate in three scenarios: 1) a 2D simulation (Section~\ref{sec:2dsim}), 2) a 3D synthetic semi-static dataset (Section~\ref{sec:3dsim}), and 3) our real-world, semi-static warehouse dataset (Section~\ref{sec:warehouse}). The lack of large, real-world SLAM datasets with multiple passes through environments that include both dynamic and semi-static objects prompted us to create one (Section~\ref{dataset}). We implement our method on top of ORB-SLAM3~\cite{campos2020orb} and POCD~\cite{QianChatrathPOCD}. The parameters used to evaluate against all methods can be found in the Supplementary Material.

To benchmark 3D reconstruction accuracy under scene changes, we generate the ground truth meshes by using POCD with the ground truth robot trajectory. As POCD has shown to outperform several mapping methods (Kimera~\cite{Rosinol20icra-Kimera}, Fehr \textit{et al.}~\cite{fehr2017tsdf}, and Panoptic Multi-TSDFs~\cite{panoptictsdf}), the mesh obtained is representative of the best possible reconstruction.

Note that in this work, we use RGB-D information to address scene changes directly, thus inertial and odometry data are excluded. While IMUs can supplement all methods, our method yields RGB-D pose estimates that align better with IMU data, removing errors at their source.

\subsection{Real-World Semi-Static Warehouse Dataset} \label{dataset}

We release an extension to the TorWIC change detection dataset~\cite{QianChatrathPOCD}. The original TorWIC dataset features a small 10m$\times$10m hallway setup using boxes and fences with limited real-world objects and changes. Its ground-truth trajectory, acquired via 2D LiDAR SLAM, suffers from jumps and drifts and thus not suitable for evaluating SLAM algorithms. Conversely, the new extension, as the first long-term real-world warehouse dataset, originates from an active 100m$\times$80m Clearpath Robotics plant showcasing various objects and changes (e.g., forklifts, robots, people). 

The dataset is collected on a mobile base equipped with two Microsoft Azure RGB-D cameras, an Ouster 128-beam LiDAR, and two IMUs. We repeat a few scenarios over the course of four months, presenting changed object locations over time. The robot setup, sensor specifications and the scenario breakdown can be found in the Supplementary Material. Figure~\ref{fig:scene_change} shows the scenario changes for a sample route.

To facilitate SLAM and reconstruction evaluation, we also release the ground truth scan of the warehouse and ground truth trajectories. A Leica MS60 multistation was used to obtain a centimetre-level accurate point cloud of the warehouse. Iterative closest point (ICP) was performed between the 128-beam LiDAR scan and the ground truth scan to obtain highly accurate ground truth trajectories for the robot. The robot starts and ends at the pre-defined map origin, so users can easily stitch trajectories to create long routes with change. 

\subsection{2D Semi-Static Simulation}
\label{sec:2dsim}

\begin{figure}[t!]
  \centering
  \includegraphics[width=0.8\columnwidth]{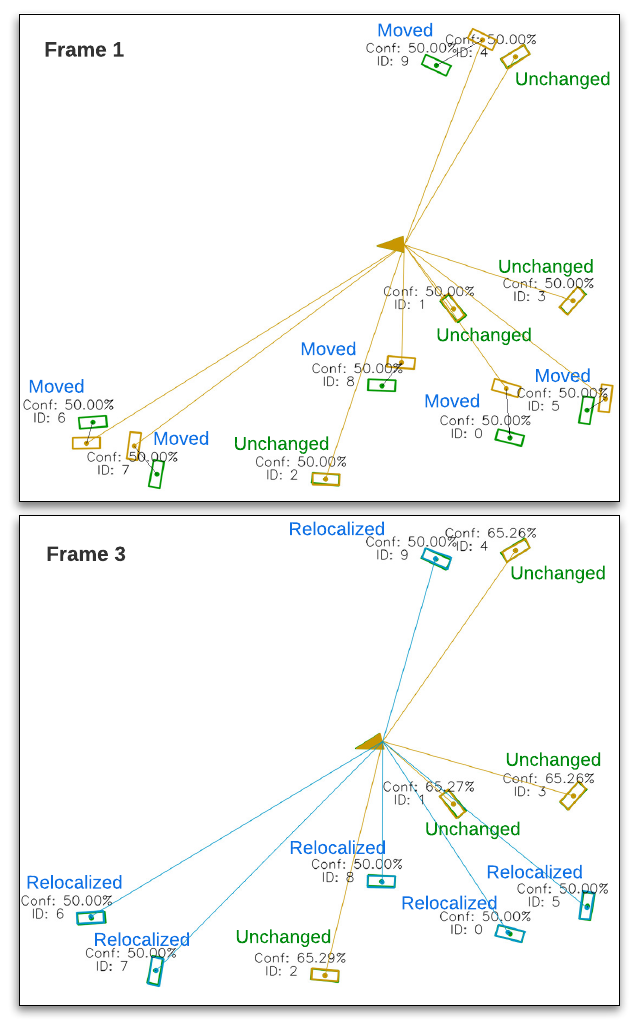}
  \caption{The setup of our 2D semi-static simulation with six moved objects and four unchanged objects over the first and third frames of optimization. \textbf{Rectangle}: object, \textbf{triangle}: robot, \textbf{green}: ground truth, \textbf{yellow}: estimation, \textbf{blue}: reconstructed objects. All object states are initialized with a $\mathbb{E}[v]=0.5$ consistency estimate. All moved boxes are identified and relocalized after three frames.}
  \label{fig:2dsetup}
  \vspace{2mm}
\end{figure}

In this section, we introduce the first of three experiments performed. A 2D simulation was constructed to demonstrate our probabilistic framework in action, and to justify the design choices made. The setup can be seen in Figure~\ref{fig:2dsetup}, consisting of four unchanged objects and six moved objects. The robot is spawned around its ground truth pose, with noise in both position and orientation, and drives in the scene. The robot measures the four vertices of the rectangular objects, all corrupted by Gaussian noise. As seen in Figure~\ref{fig:2dsetup}, the system is able to correctly identify the six moved boxes, recovering their true poses. In the Supplementary Material, the evolution of the state estimates of the system over the first four frames are shown, as the robot navigates the scene.

\begin{figure}[t!]
  \centering
  \includegraphics[width=0.8\columnwidth]{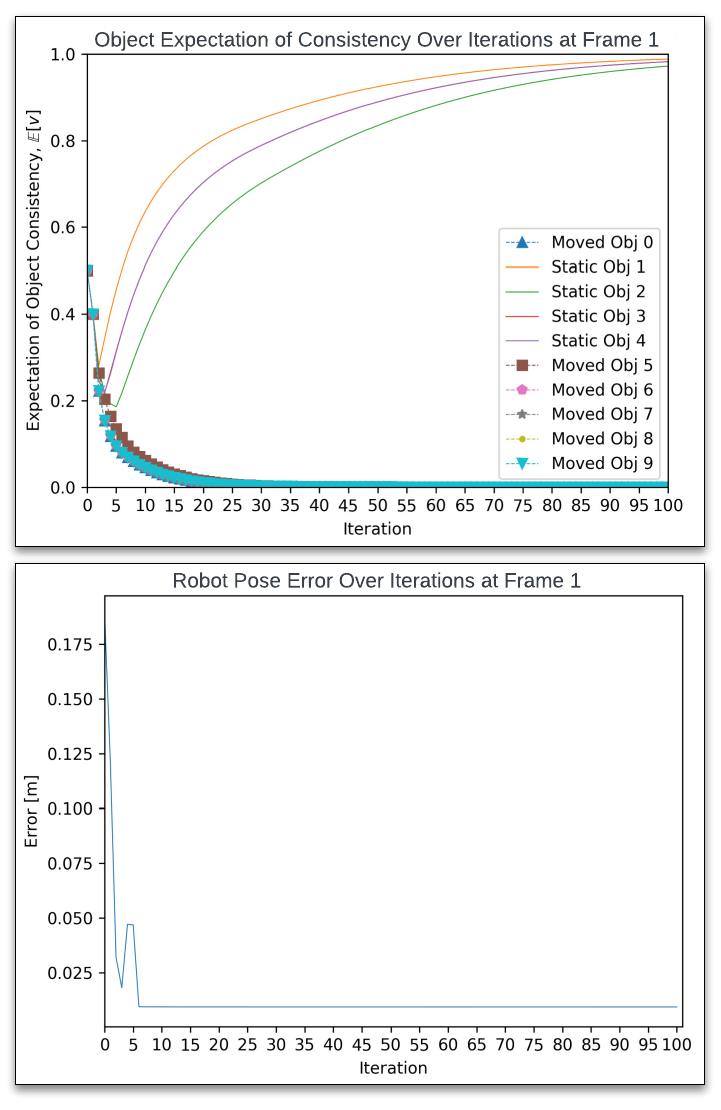}
  \caption{The object-level consistency expectation (top) and robot pose error (bottom) over VEM iterations at the first frame for the simulated scenario. For the object consistencies, plots with markers belong to moved objects. Although the consistencies converge after 100 iterations, in practice we can stop much earlier.}
  \label{fig:compare_charts}
\end{figure}

Figure~\ref{fig:compare_charts} shows the evolution of the object consistency expectation, $\mathbb{E}[v]$, and the robot pose error over the EM iterations at the first frame. The consistency expectations converge to their true values at the end of the optimization and the robot pose converges to its ground truth after six EM iterations. There is a drop in the consistency of all objects during the first iterations due to the initial error in the robot pose estimate. However, as the robot pose becomes more accurate, the true states are recovered. This experiment shows the robustness of our method, as the system is able to recover the true state even when the number of moved objects in the scene exceeds the number of unchanged objects. 

We shall note that the iterative optimization process finds the most likely underlying scene configuration based on the measurements. Therefore, if there exists a different hypothesis that exhibits a higher measurement likelihood, the optimizer would converge to that solution. For example, if the six moved objects had all shifted in the same direction by the same magnitude, our system would mark them as stationary and relocalize the unchanged objects instead. However, since such scenarios cannot be distinguished from a probabilistic point of view, they are not of concern. This adversarial scenario is shown in the Supplementary Material. 

Further ablation studies showcasing the advantage of using the ELBO instead of the single point estimate (Section~\ref{sec:factor_graph_overview}), the use of max-mixture approximation (Section~\ref{sec:max_mix}), the choice of weights ($\mathbb{E}[v]$ vs $\mathbb{E}[\pi]$) to use when choosing the mode of max-mixture (Section~\ref{sec:max_mix}), and an adversarial fully dynamic scenario, are available in the Supplementary Material. 

\begin{figure}[t!]
  \centering
  \includegraphics[width=\columnwidth]{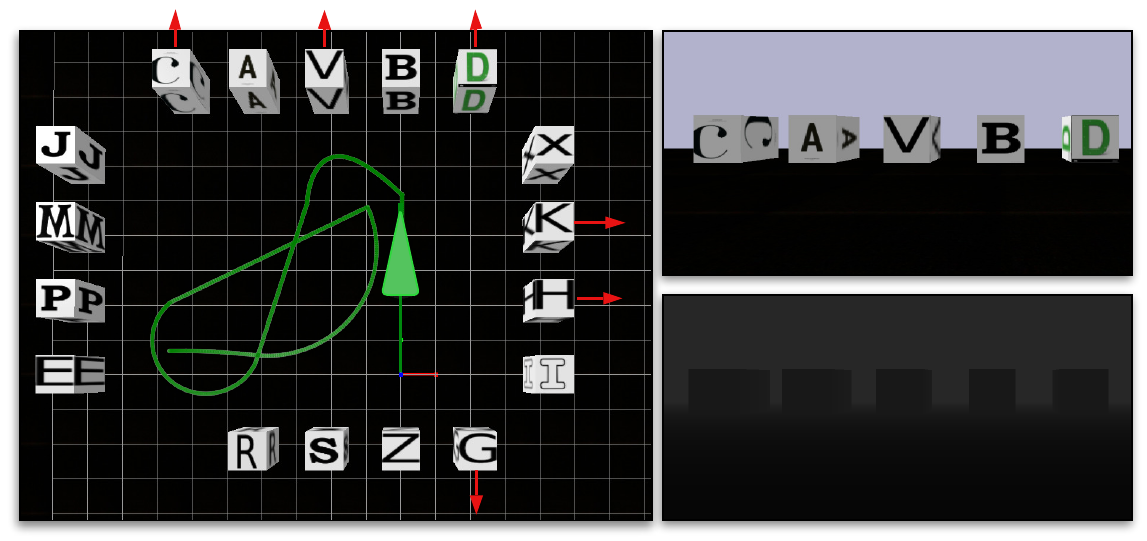}
  \caption{The setup for the 3D simulation (left). \textbf{Green}: robot trajectory, \textbf{red}: direction of box motion. The RGB (top) and depth (bottom) camera views can be seen on the right.}
  \label{fig:3dsetup}
\end{figure}

\begin{figure*}[t!]
  \centering
  \includegraphics[scale=0.8]{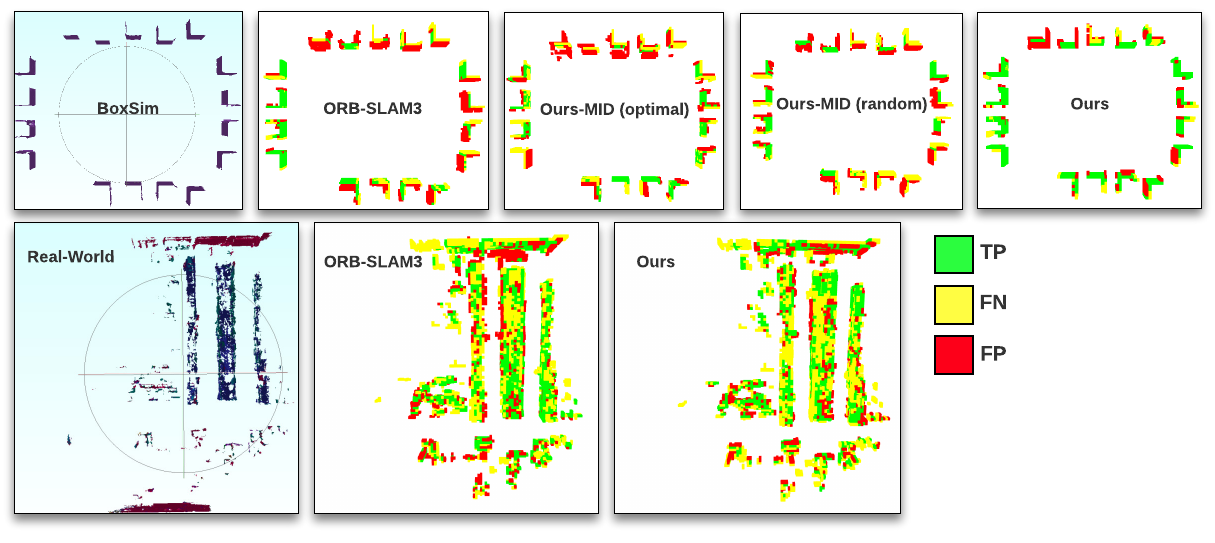}
  \caption{A bird's-eye-view qualitative analysis of 3D reconstruction results of the \textbf{top row:} BoxSim synthetic dataset compared against that of ORB-SLAM3 and Ours-MID, and \textbf{bottom row:} our real-world warehouse dataset, compared against that of ORB-SLAM3. The green, yellow, and red sections represent true positives (correct prediction), false negatives (incorrect negative prediction), and false positives (incorrect prediction), respectively. The first image is the ground truth map of the routes' final scenario after scene change. For BoxSims a grid of 0.2 and for the real-world trajectory a grid size of 0.4 were used to voxelize the reconstruction.}
  \label{fig:results_3dsim}
\end{figure*}

\subsection{3D Semi-Static Simulation}
\label{sec:3dsim}

In this section, we introduce a 3D simulated semi-static scene, henceforth referred to as \say{BoxSim}. The setup can be seen in Figure~\ref{fig:3dsetup}. The robot moves among 17 boxes, six of which shift between robot traversals. The scenario is very challenging as there is no static background available, requiring all methods to localize against the 17 boxes.

Figure~\ref{fig:fig1} visually compares the robot trajectories estimated by ORB-SLAM3 and \method~against the ground truth. As discussed, ORB-SLAM3 assumes a static environment. Although it utilizes robust kernels and iterative pruning to reject outlier landmarks, it is still sensitive to large scene change. As seen in the figure, its estimated trajectory diverges from the ground truth when changed objects are encountered. On the other hand,~\method~optimizes a lower bound to a Gaussian-Uniform likelihood to explicitly infer if any of the mapped objects have changed, resulting in much smoother and accurate pose estimates.

\begin{table}[b!]
%\fontsize{7pt}{5pt}\selectfont
\vspace{2mm}
\centering
\caption{Absolute Trajectory Error (ATE) and Maximum Position Error (MPE) on the BoxSim Dataset.}
\scriptsize
%\resizebox{\columnwidth}{!}{
\begin{tabular}{c|cc}
\hline
\Xhline{2\arrayrulewidth}  \Xhline{2\arrayrulewidth}
\textbf{BoxSim} & \textbf{ATE [m]} & \textbf{MPE [m]}  \\ \hline
ORB-SLAM3   &   0.14 & 0.47  \\
Ours-MID (optimal)    &  0.41  & 0.82 \\
Ours-MID (random)    &  0.49 &  0.90 \\
\hline 
\textbf{\method~(ours)} &  \textbf{0.10} & \textbf{0.26} \\ 
\rowcolor{LightCyan}
\textit{\method~ Improvement} & 0.04 (29\%) & 0.21  (45\%)\\ 
\hline
\end{tabular}%}
\label{tab:3dsim}
\end{table}

\begin{table}[b!]
\vspace{2mm}
\centering
\caption{Quantitative mapping results on the BoxSim Dataset.}
\scriptsize
%\resizebox{\columnwidth}{!}{
\begin{tabular}{c|ccc}
\hline
\Xhline{2\arrayrulewidth}  \Xhline{2\arrayrulewidth}

\textbf{BoxSim} & \textbf{Precision $\uparrow$} & \textbf{Recall (TPR) $\uparrow$} & \textbf{FPR $\downarrow$} \\ \hline

ORB-SLAM3   &  52.1 &  72.4 & 3.8  \\
Ours-MID (optimal)    &  45.2 & 56.7 & 4.1 \\
Ours-MID (random)    &  39.0 & 50.5 & 4.6 \\

\hline 
\textbf{\method~(ours)}    & \textbf{68.5}  & \textbf{78.7} & \textbf{2.2} \\ 
\rowcolor{LightCyan}
\textit{\method~ Improvement} & 16.4  (31\%) & 6.3 (8.7\%) & -1.6 (42\%)\\ 
\hline
\end{tabular}%}
\label{tab:3dsimQuant}
\end{table}

As discussed in the literature review, a common method for dynamic object handling is to ignore all potentially moving objects. In VI-MID~\cite{visinsMulti} the authors mask out all potentially-changing objects based on semantic information, performing dense tracking on the static background alone. However, this relies on the assumption that the changing parts of the environment are known, which is not feasible in the real world. We evaluate our adaptation, Ours-MID, on two scenarios: 1) the \textbf{optimal} case, where the system knows which objects will shift and 2) the \textbf{random} case, where objects are randomly chosen to represent the static background.

The average trajectory error (ATE) and maximum position error (MPE) can be seen in Table~\ref{tab:3dsim}. \method~significantly outperforms ORB-SLAM3 and Ours-MID. For Ours-MID, in the \textbf{optimal} case, by ignoring all potentially changing objects the system might not observe enough features when the robot visits locations where moving objects dominate, causing poor estimates. In the \textbf{random} case, its performance further degrades when objects are incorrectly classified.

We further compare the dense reconstructions of \method~against ORB-SLAM3 and Ours-MID. The top row of Figure~\ref{fig:results_3dsim} shows the qualitative comparisons, where we overlay and voxelize both the reconstructions and the ground truth mesh and colorize the overlapping (inlier) and inconsistent (outlier) voxels. We then compute the precision, recall, and false positive rate (FPR) by counting the voxels for a quantitative evaluation, and Table~\ref{tab:3dsimQuant} lists the quantitative results. ORB-SLAM3 and Ours-MID both generated distorted maps with failed object updates due to localization drift, which led to incorrect data association in object consistency update. On the other hand, \method~generates the most visually correct map where all moved boxes, except the one at the top left, are relocalized to the new locations. Quantitatively, \method~exhibits the highest precision (coverage of true objects), and the lowest FPR (map update quality after scene change) due to its superior localization performance. 

As all methods use the same POCD~\cite{QianChatrathPOCD} framework and parameters to perform map update at every frame, this experiment highlights that 1) explicit reasoning of object consistency is required for localization in semi-static environments, and 2) joint estimation of object consistency and robot localization brings significant advantage in cluttered scenes.

\subsection{Real-World Experiment in a Semi-Static Scene}
\label{sec:warehouse}

\begin{figure}[t!]
  \centering
  \includegraphics[width=0.7\columnwidth]{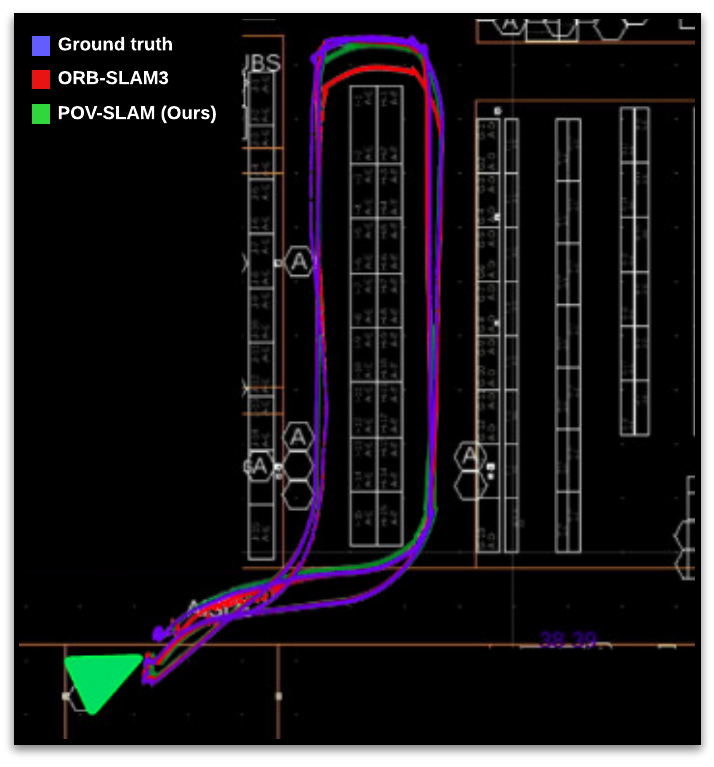}
  \caption{The output trajectories of ORB-SLAM3 and~\method~through the real-world warehouse aisle scenario.}
  \label{fig:warehouse_traj}
\end{figure}

In this section, we evaluate \method's~effectiveness in a warehouse scenario, available through our novel real-world semi-static dataset. We stitch two trajectories captured along the same route four months apart to introduce scene changes as the robot traverses the warehouse. Figure~\ref{fig:warehouse_traj} shows the routes overlaid on the factory's schematic floor plan and Figure~\ref{fig:scene_change} shows a sample pair of frames with scene changes. Due to the limited effective range of the Azure RGB-D cameras, we rely on the Ouster Lidar to provide feature depth information when traversing in open areas. 

We qualitatively and quantitatively compare the trajectory estimation and scene reconstruction results against ORB-SLAM3. Ours-MID is not included in the comparison as the route is cluttered with pallets and boxes, leaving very limited static background information for Ours-MID to localize against. The output trajectories along with the ground truth are visualized in Figure~\ref{fig:warehouse_traj}. ORB-SLAM3 successfully completes the first traversal with high accuracy. However, changes along the aisle in the second traversal cause incorrect data association and lead to a shortened trajectory. \method~performs slightly worse than ORB-SLAM3 in the first traversal. However, in the second traversal, \method~is able to reject the false positive matches and track with higher accuracy. The ATEs and MPEs from the two traversals can be seen in Table~\ref{tab:warehouse1} and Table~\ref{tab:warehouse2}.

The bottom row of Figure~\ref{fig:results_3dsim} visualizes the 3D reconstruction results. Again, we voxelize the reconstructed meshes and count for overlapping and inconsistent voxels to obtain the quantitative evaluations, which are listed in Table~\ref{tab:warehousQuant}. \method~outperforms ORB-SLAM3 in both localization accuracy and scene reconstruction on this route when scene changes are encountered as it does not suffer from incorrect loop closures.

\begin{table}[t!]
\vspace{5mm}
\centering
\caption{Absolute Trajectory Error (ATE) and Maximum Position Error (MPE) in Traversal 1 of the Real-World Aisle Scenario.}
\scriptsize
%\resizebox{\columnwidth}{!}{
\begin{tabular}{c|cc}
\hline
\Xhline{2\arrayrulewidth}  \Xhline{2\arrayrulewidth}

\textbf{Real-World-T1} & \textbf{ATE-T1 [m]} & \textbf{MPE-T1 [m]} \\ \hline
ORB-SLAM3   &   \textbf{0.14} & \textbf{0.31}   \\
\hline 
\textbf{\method~(ours)} &  0.26 & 0.48    \\ 
\rowcolor{LightCyan}
\textit{\method~ Improvement} & -0.12 (-86\%) & -0.17 (-55\%) \\ 
\hline
\end{tabular}%}
\label{tab:warehouse1}
\end{table}

\begin{table}[t!]
\vspace{5mm}
\centering
\caption{Absolute Trajectory Error (ATE) and Maximum Position Error (MPE) in Traversal 2 of the Real-World Aisle Scenario.}
\scriptsize
%\resizebox{\columnwidth}{!}{
\begin{tabular}{c|cc}
\hline
\Xhline{2\arrayrulewidth}  \Xhline{2\arrayrulewidth}

\textbf{Real-World-T2} &  \textbf{ATE-T2 [m]} & \textbf{MPE-T2 [m]} \\ \hline
ORB-SLAM3    &   0.89 & 1.55 \\
\hline 
\textbf{\method~(ours)} &  \textbf{0.46} & \textbf{0.98} \\ 
\rowcolor{LightCyan}
\textit{\method~ Improvement}  & 0.43 (48\%) & 0.57 (37\%)\\ 
\hline
\end{tabular}%}
\label{tab:warehouse2}
\end{table}

\begin{table}[t!]
\scriptsize
\vspace{2mm}
\centering
\caption{Quantitative mapping results on the Real World Dataset.}
%\resizebox{\columnwidth}{!}{
\begin{tabular}{c|ccc}
\hline
\Xhline{2\arrayrulewidth}  \Xhline{2\arrayrulewidth}

\textbf{Real World} & \textbf{Precision $\uparrow$} & \textbf{Recall (TPR) $\uparrow$} & \textbf{FPR $\downarrow$} \\ \hline
ORB-SLAM3   & 76.2 & \textbf{56.9} & 1.9 \\

\hline 
\textbf{\method~(ours)}    & \textbf{79.7}  & 53.7 & \textbf{1.5} \\
\rowcolor{LightCyan}
\textit{\method~ Improvement} & 3.5  (4.6\%)& -3.2 (-6.0\%) & -0.4 (21\%) \\ 
\hline
\end{tabular}%}
\label{tab:warehousQuant}
\end{table}

\subsection{Run-time Performance}
With a max of 4,000 ORB features in each frame, a window size of 8, and 30 EM iterations per frame, \method~runs at approximately 1Hz on a Linux desktop with an AMD Ryzen R9-5900X CPU at 3.7Hz. To achieve a more realistic run-time, we only execute the VEM optimization every seven frames on the real-world dataset, while relying on ORB-SLAM3 in between. We use a large number of ORB features because the dataset is challenging due to varying lighting conditions, causing even the original ORB-SLAM3 to fail at times with the default 1250 features. As well, our object-aware method requires each object to have sufficient features for tracking and association. We currently use a uniform feature detection approach, so small yet key objects may not get enough features under a lower quota. In practice, \method~is amenable to online operation in large environments, as change detection and localization correction is not required at every frame. A semantic-aware feature extraction approach could further improve the performance in the future.

%%%%%% Conclusion
\section{Conclusion}\label{sec:conclusion}
In this paper we present \method, a novel online, probabilistic object-aware framework to simultaneously estimate the robot pose, and track and update object-level scene changes in a joint optimization-based framework. The~\method~pipeline uses our derived variational expectation maximization strategy to optimize factor graphs accounting for potentially-changing objects. We experimentally verify the robustness of \method~against state-of-the-art SLAM methods on two datasets, including our novel, real-world, semi-static warehouse dataset that we release with this work. Our system explicitly reasons about object-level stationarity to improve the robustness of localization in slowly varying scenes. Our method outperforms ORB-SLAM3 on average trajectory error by 48\% on the real-world dataset and 29\% on the 3D synthetic semi-static dataset. As well, \method~shows a 4.6\% improvement on dense reconstruction precision in the large real-world scene and 31\% in the smaller synthetic scene.

\clearpage
% We plan to extend our current framework in several ways. First, our naive data association suffers from partial observations, which we aim to address by using the recently proposed latent object model (e.g., DeepSDF \cite{deepsdf}). Second, we can improve the efficiency of how dynamic objects are handled by performing explicit motion tracking, as done in \cite{Rnz2018MaskFusionRR}. Third, we will explore the effect of leveraging the object-level geometric and stationarity information on state estimation in dynamic scenes. Finally, as the robot can be faced with multimodal uncertainties in real-world deployment, we aim to explore expressive multimodal representations to improve robustness against noisy measurements.

%% Use plainnat to work nicely with natbib. 

\bibliographystyle{unsrtnat}
\bibliography{references}

\begin{thebibliography}{42}
\providecommand{\natexlab}[1]{#1}
\providecommand{\url}[1]{\texttt{#1}}
\expandafter\ifx\csname urlstyle\endcsname\relax
  \providecommand{\doi}[1]{doi: #1}\else
  \providecommand{\doi}{doi: \begingroup \urlstyle{rm}\Url}\fi

\bibitem[Mur-Artal et~al.(2015)Mur-Artal, Montiel, and Tardos]{mur2015orb}
Raul Mur-Artal, Jose Maria~Martinez Montiel, and Juan~D. Tardos.
\newblock {ORB-SLAM}: a versatile and accurate monocular {SLAM} system.
\newblock \emph{IEEE Transactions on Robotics}, 2015.

\bibitem[{Grinvald} et~al.(2019){Grinvald}, {Furrer}, {Novkovic}, {Chung},
  {Cadena}, {Siegwart}, and {Nieto}]{grinvald2019volumetric}
M.~{Grinvald}, F.~{Furrer}, T.~{Novkovic}, J.~J. {Chung}, C.~{Cadena},
  R.~{Siegwart}, and J.~{Nieto}.
\newblock {Volumetric Instance-Aware Semantic Mapping and 3D Object Discovery}.
\newblock \emph{IEEE Robotics and Automation Letters}, 2019.

\bibitem[Rosinol et~al.(2020{\natexlab{a}})Rosinol, Abate, Chang, and
  Carlone]{Rosinol20icra-Kimera}
Antoni Rosinol, Marcus Abate, Yun Chang, and Luca Carlone.
\newblock Kimera: an open-source library for real-time metric-semantic
  localization and mapping.
\newblock In \emph{2020 IEEE International Conference on Robotics and
  Automation (ICRA)}, 2020{\natexlab{a}}.

\bibitem[Whelan et~al.(2016)Whelan, Salas-Moreno, Glocker, Davison, and
  Leutenegger]{whelan2016elasticfusion}
Thomas Whelan, Renato~F Salas-Moreno, Ben Glocker, Andrew~J Davison, and Stefan
  Leutenegger.
\newblock Elasticfusion: Real-time dense {SLAM} and light source estimation.
\newblock \emph{The International Journal of Robotics Research}, 2016.

\bibitem[Campos et~al.(2020)Campos, Elvira, Rodr{\'\i}guez, Montiel, and
  Tard{\'o}s]{campos2020orb}
Carlos Campos, Richard Elvira, Juan J~G{\'o}mez Rodr{\'\i}guez, Jos{\'e}~MM
  Montiel, and Juan~D Tard{\'o}s.
\newblock {ORB-SLAM}3: An accurate open-source library for visual,
  visual-inertial and multi-map {SLAM}.
\newblock \emph{arXiv preprint arXiv:2007.11898}, 2020.

\bibitem[Qian et~al.(2022)Qian, Chatrath, Yang, Servos, Schoellig, and
  Waslander]{QianChatrathPOCD}
Jingxing Qian, Veronica Chatrath, Jun Yang, James Servos, Angela Schoellig, and
  Steven~L. Waslander.
\newblock {POCD: Probabilistic Object-Level Change Detection and Volumetric
  Mapping in Semi-Static Scenes}.
\newblock In \emph{2022 Robotics: Science and Systems (RSS)}, 2022.

\bibitem[Ballester et~al.(2020)Ballester, Font{\'{a}}n, Civera, Strobl, and
  Triebel]{DOT}
Irene Ballester, Alejandro Font{\'{a}}n, Javier Civera, Klaus~H. Strobl, and
  Rudolph Triebel.
\newblock {DOT:} dynamic object tracking for visual {SLAM}.
\newblock \emph{CoRR}, 2020.

\bibitem[Yu et~al.(2018)Yu, Liu, Liu, Xie, Yang, Wei, and Qiao]{DS_SLAM}
Chao Yu, Zuxin Liu, Xinjun Liu, Fugui Xie, Yi~Yang, Qi~Wei, and Fei Qiao.
\newblock {DS-SLAM:} {A} semantic visual {SLAM} towards dynamic environments.
\newblock \emph{CoRR}, 2018.

\bibitem[Lu et~al.(2020)Lu, Wang, Tang, Huang, and Li]{DM-SLAM}
Xiaoyun Lu, Hu~Wang, Shuming Tang, Huimin Huang, and Chuang Li.
\newblock Dm-slam: Monocular {SLAM} in dynamic environments.
\newblock \emph{Applied Sciences}, 2020.

\bibitem[McCormac et~al.(2018)McCormac, Clark, Bloesch, Davison, and
  Leutenegger]{fusion++}
J.~McCormac, R.~Clark, Michael Bloesch, A.~Davison, and Stefan Leutenegger.
\newblock Fusion++: Volumetric object-level {SLAM}.
\newblock \emph{2018 International Conference on 3D Vision (3DV)}, 2018.

\bibitem[Sun et~al.(2019)Sun, Sun, Liu, and Yeung]{Sun2019MovableObjectAwareVS}
Ting Sun, Yuxiang Sun, Ming Liu, and Dit-Yan Yeung.
\newblock Movable-object-aware visual {SLAM} via weakly supervised semantic
  segmentation.
\newblock \emph{ArXiv}, 2019.

\bibitem[Rosinol et~al.(2020{\natexlab{b}})Rosinol, Abate, Shi, and
  Carlone]{dsg}
Antoni Rosinol, Marcus Abate, Jingnan Shi, and Luca Carlone.
\newblock 3d dynamic scene graphs: Actionable spatial perception with places,
  objects, and humans.
\newblock In \emph{Robotics: Science and Systems}, 2020{\natexlab{b}}.

\bibitem[R{\"u}nz and de~Agapito(2018)]{Rnz2018MaskFusionRR}
Martin R{\"u}nz and Lourdes de~Agapito.
\newblock Maskfusion: Real-time recognition, tracking and reconstruction of
  multiple moving objects.
\newblock \emph{2018 IEEE International Symposium on Mixed and Augmented
  Reality (ISMAR)}, 2018.

\bibitem[Hachiuma et~al.(2019)Hachiuma, Pirchheim, Schmalstieg, and
  Saito]{Hachiuma2019DetectFusionDA}
Ryo Hachiuma, Christian Pirchheim, Dieter Schmalstieg, and H.~Saito.
\newblock Detectfusion: Detecting and segmenting both known and unknown dynamic
  objects in real-time {SLAM}.
\newblock In \emph{BMVC}, 2019.

\bibitem[Xu et~al.(2019)Xu, Li, Tzoumanikas, Bloesch, Davison, and
  Leutenegger]{Xu2019MIDFusionOO}
Binbin Xu, Wenbin Li, Dimos Tzoumanikas, Michael Bloesch, Andrew~J. Davison,
  and Stefan Leutenegger.
\newblock {MID}-fusion: Octree-based object-level multi-instance dynamic
  {SLAM}.
\newblock \emph{2019 International Conference on Robotics and Automation
  (ICRA)}, 2019.

\bibitem[Wang et~al.(2021)Wang, Luo, Zhang, Zhao, Yin, Wang, Su, Wang, and
  Li]{Wang2021DymSLAM4D}
Chenjie Wang, Bin Luo, Yun Zhang, Qing Zhao, Lu-Jun Yin, Wei Wang, Xin Su,
  Yajun Wang, and Chengyuan Li.
\newblock {DymSLAM}: 4d dynamic scene reconstruction based on geometrical
  motion segmentation.
\newblock \emph{IEEE Robotics and Automation Letters}, 2021.

\bibitem[Ren et~al.(2022)Ren, Xu, Choi, and Leutenegger]{visinsMulti}
Yifei Ren, Binbin Xu, Christopher~L. Choi, and Stefan Leutenegger.
\newblock Visual-inertial multi-instance dynamic {SLAM} with object-level
  relocalisation.
\newblock In \emph{arXiv}, 2022.

\bibitem[Fehr et~al.(2017)Fehr, Furrer, Dryanovski, Sturm, Gilitschenski,
  Siegwart, and Cadena]{fehr2017tsdf}
Marius Fehr, Fadri Furrer, Ivan Dryanovski, J{\"u}rgen Sturm, Igor
  Gilitschenski, Roland Siegwart, and Cesar Cadena.
\newblock {TSDF}-based change detection for consistent long-term dense
  reconstruction and dynamic object discovery.
\newblock In \emph{2017 IEEE International Conference on Robotics and
  automation (ICRA)}. IEEE, 2017.

\bibitem[Schmid et~al.(2021)Schmid, Delmerico, Sch{\"{o}}nberger, Nieto,
  Pollefeys, Siegwart, and Cadena]{panoptictsdf}
Lukas~Maximilian Schmid, Jeffrey~A. Delmerico, Johannes~L. Sch{\"{o}}nberger,
  Juan~I. Nieto, Marc Pollefeys, Roland Siegwart, and C{\'{e}}sar Cadena.
\newblock Panoptic multi-{TSDF}s: a flexible representation for online
  multi-resolution volumetric mapping and long-term dynamic scene consistency.
\newblock \emph{CoRR}, 2021.

\bibitem[Ahn et~al.(2022)Ahn, Brohan, Brown, Chebotar, Cortes, David, Finn, Fu,
  Gopalakrishnan, Hausman, Herzog, Ho, Hsu, Ibarz, Ichter, Irpan, Jang, Ruano,
  Jeffrey, Jesmonth, Joshi, Julian, Kalashnikov, Kuang, Lee, Levine, Lu, Luu,
  Parada, Pastor, Quiambao, Rao, Rettinghouse, Reyes, Sermanet, Sievers, Tan,
  Toshev, Vanhoucke, Xia, Xiao, Xu, Xu, Yan, and Zeng]{doasican}
Michael Ahn, Anthony Brohan, Noah Brown, Yevgen Chebotar, Omar Cortes, Byron
  David, Chelsea Finn, Chuyuan Fu, Keerthana Gopalakrishnan, Karol Hausman,
  Alex Herzog, Daniel Ho, Jasmine Hsu, Julian Ibarz, Brian Ichter, Alex Irpan,
  Eric Jang, Rosario~Jauregui Ruano, Kyle Jeffrey, Sally Jesmonth, Nikhil
  Joshi, Ryan Julian, Dmitry Kalashnikov, Yuheng Kuang, Kuang-Huei Lee, Sergey
  Levine, Yao Lu, Linda Luu, Carolina Parada, Peter Pastor, Jornell Quiambao,
  Kanishka Rao, Jarek Rettinghouse, Diego Reyes, Pierre Sermanet, Nicolas
  Sievers, Clayton Tan, Alexander Toshev, Vincent Vanhoucke, Fei Xia, Ted Xiao,
  Peng Xu, Sichun Xu, Mengyuan Yan, and Andy Zeng.
\newblock Do as i can and not as i say: Grounding language in robotic
  affordances.
\newblock In \emph{arXiv preprint arXiv:2204.01691}, 2022.

\bibitem[Zhou et~al.(2022)Zhou, Wen, Wang, Gao, Li, Wang, Yang, Lu, Cao, Xu,
  and Gao]{swarm_microrobots}
Xin Zhou, Xiangyong Wen, Zhepei Wang, Yuman Gao, Haojia Li, Qianhao Wang,
  Tiankai Yang, Haojian Lu, Yanjun Cao, Chao Xu, and Fei Gao.
\newblock Swarm of micro flying robots in the wild.
\newblock \emph{Science Robotics}, 2022.

\bibitem[Klein and Murray(2007)]{klein2007parallel}
Georg Klein and David Murray.
\newblock Parallel tracking and mapping for small ar workspaces.
\newblock In \emph{2007 6th IEEE and ACM international symposium on mixed and
  augmented reality}. IEEE, 2007.

\bibitem[Endres et~al.(2014)Endres, Hess, Sturm, Cremers, and
  Burgard]{rgbdslam}
Felix Endres, Jurgen Hess, Jurgen Sturm, Daniel Cremers, and Wolfram Burgard.
\newblock 3-d mapping with an {RGB-D} camera.
\newblock \emph{Robotics, IEEE Transactions on}, 2014.

\bibitem[Gridseth and Barfoot(2019)]{vtar}
Mona Gridseth and Timothy Barfoot.
\newblock Towards direct localization for visual teach and repeat.
\newblock In \emph{2019 16th Conference on Computer and Robot Vision (CRV)},
  2019.

\bibitem[Newcombe et~al.(2011)Newcombe, Izadi, Hilliges, Molyneaux, Kim,
  Davison, Kohi, Shotton, Hodges, and Fitzgibbon]{newcombe2011kinectfusion}
Richard~A Newcombe, Shahram Izadi, Otmar Hilliges, David Molyneaux, David Kim,
  Andrew~J Davison, Pushmeet Kohi, Jamie Shotton, Steve Hodges, and Andrew
  Fitzgibbon.
\newblock Kinectfusion: Real-time dense surface mapping and tracking.
\newblock In \emph{2011 10th IEEE International Symposium on Mixed and
  Augmented Reality}. IEEE, 2011.

\bibitem[Mur-Artal and Tard{\'o}s(2017)]{mur2017orb}
Raul Mur-Artal and Juan~D Tard{\'o}s.
\newblock {ORB-SLAM}2: An open-source {SLAM} system for monocular, stereo, and
  {RGB-D} cameras.
\newblock \emph{IEEE Transactions on Robotics}, 2017.

\bibitem[Servières et~al.(2021)Servières, Renaudin, Dupuis, and
  Antigny]{survey-slam}
Myriam Servières, Valérie Renaudin, Alexis Dupuis, and Nicolas Antigny.
\newblock Visual and visual-inertial {SLAM}: State of the art, classification,
  and experimental benchmarking.
\newblock \emph{Journal of Sensors}, 2021.

\bibitem[Gao et~al.(2020)Gao, Lang, and Ren]{survey2-stereovis}
Boyu Gao, Haoxiang Lang, and Jing Ren.
\newblock Stereo visual {SLAM} for autonomous vehicles: A review.
\newblock In \emph{2020 IEEE International Conference on Systems, Man, and
  Cybernetics (SMC)}, 2020.

\bibitem[{He} et~al.(2017){He}, {Gkioxari}, {Dollár}, and
  {Girshick}]{maskrcnn}
K.~{He}, G.~{Gkioxari}, P.~{Dollár}, and R.~{Girshick}.
\newblock Mask r-cnn.
\newblock In \emph{2017 IEEE International Conference on Computer Vision
  (ICCV)}, 2017.

\bibitem[Ganti and Waslander(2019)]{SIVO}
Pranav Ganti and Steven~L. Waslander.
\newblock Network uncertainty informed semantic feature selection for visual
  slam.
\newblock In \emph{2019 16th Conference on Computer and Robot Vision (CRV)},
  2019.

\bibitem[B{\^{a}}rsan et~al.(2019)B{\^{a}}rsan, Liu, Pollefeys, and
  Geiger]{dynslam}
Ioan~Andrei B{\^{a}}rsan, Peidong Liu, Marc Pollefeys, and Andreas Geiger.
\newblock Robust dense mapping for large-scale dynamic environments.
\newblock \emph{CoRR}, abs/1905.02781, 2019.
\newblock URL \url{http://arxiv.org/abs/1905.02781}.

\bibitem[Strecke and St{\"u}ckler(2019)]{strecke2019_emfusion}
Michael Strecke and J{\"o}rg St{\"u}ckler.
\newblock {EM}-fusion: Dynamic object-level {SLAM} with probabilistic data
  association.
\newblock In \emph{Proceedings IEEE/CVF International Conference on Computer
  Vision 2019 (ICCV)}. IEEE, 2019.

\bibitem[Zhang et~al.(2020)Zhang, Henein, Mahony, and Ila]{zhang2020vdoslam}
Jun Zhang, Mina Henein, Robert Mahony, and Viorela Ila.
\newblock {VDO-SLAM: A Visual Dynamic Object-aware {SLAM} System}.
\newblock \emph{arXiv}, 2020.

\bibitem[Huang et~al.(2021)Huang, Yang, Zhao, Lai, and Hu]{clusterslam}
Jiahui Huang, Sheng Yang, Zishuo Zhao, Yu-Kun Lai, and Shi-Min Hu.
\newblock Clusterslam: A {SLAM} backend for simultaneous rigid body clustering
  and motion estimation.
\newblock \emph{Computational Visual Media}, 2021.

\bibitem[Gomez et~al.(2020)Gomez, Hernandez, Derner, Barber, and
  Babuška]{object_pose_graph}
Clara Gomez, Alejandra~C. Hernandez, Erik Derner, Ramon Barber, and Robert
  Babuška.
\newblock Object-based pose graph for dynamic indoor environments.
\newblock \emph{IEEE Robotics and Automation Letters}, 2020.

\bibitem[Walcott-Bryant et~al.(2012)Walcott-Bryant, Kaess, Johannsson, and
  Leonard]{walcott2012dynamic}
Aisha Walcott-Bryant, Michael Kaess, Hordur Johannsson, and John~J Leonard.
\newblock Dynamic pose graph {SLAM}: Long-term mapping in low dynamic
  environments.
\newblock In \emph{2012 IEEE/RSJ International Conference on Intelligent Robots
  and Systems}. IEEE, 2012.

\bibitem[Rosen et~al.(2016)Rosen, Mason, and Leonard]{Rosen2016TowardsLF}
David~M. Rosen, Julian Mason, and John~J. Leonard.
\newblock Towards lifelong feature-based mapping in semi-static environments.
\newblock \emph{2016 IEEE International Conference on Robotics and Automation
  (ICRA)}, pages 1063--1070, 2016.

\bibitem[Rogers et~al.(2010)Rogers, Trevor, Nieto-Granda, and
  Christensen]{SLAM-EM}
John~G. Rogers, Alexander J.~B. Trevor, Carlos Nieto-Granda, and Henrik~I.
  Christensen.
\newblock {SLAM} with expectation maximization for moveable object tracking.
\newblock In \emph{2010 IEEE/RSJ International Conference on Intelligent Robots
  and Systems (IROS)}. IEEE, 2010.

\bibitem[Xiang et~al.(2015)Xiang, Ren, Ni, and Jenkins]{RobustGraphSLAM}
Lingzhu Xiang, Zhile Ren, Mengrui Ni, and Odest~Chadwicke Jenkins.
\newblock Robust graph {SLAM} in dynamic environments with moving landmarks.
\newblock In \emph{2015 IEEE/RSJ International Conference on Intelligent Robots
  and Systems (IROS)}. IEEE, 2015.

\bibitem[Murphy(2012)]{murphy2012machine}
Kevin~P Murphy.
\newblock \emph{Machine learning: a probabilistic perspective}.
\newblock MIT press, 2012.

\bibitem[Kümmerle et~al.(2011)Kümmerle, Grisetti, Strasdat, Konolige, and
  Burgard]{g20_pub}
Rainer Kümmerle, Giorgio Grisetti, Hauke Strasdat, Kurt Konolige, and Wolfram
  Burgard.
\newblock G2o: A general framework for graph optimization.
\newblock In \emph{2011 IEEE International Conference on Robotics and
  Automation}, 2011.

\bibitem[Olson and Agarwal(2012)]{Olson-max-mixture}
Edwin Olson and Pratik Agarwal.
\newblock Inference on networks of mixtures for robust robot mapping.
\newblock In \emph{Proceedings of Robotics: Science and Systems}, 2012.

\end{thebibliography}

\end{document}